\documentclass[10pt,twocolumn,letterpaper]{article}

\usepackage[pagenumbers]{cvpr}      
\definecolor{cvprblue}{rgb}{0.21,0.49,0.74}
\usepackage[pagebackref,breaklinks,colorlinks,allcolors=cvprblue]{hyperref}

\usepackage{url}
\usepackage{booktabs}
\usepackage{multicol}
\usepackage{multirow}
\usepackage{amssymb}
\usepackage{pifont}
\usepackage{float}
\usepackage[ruled,vlined]{algorithm2e}

\usepackage{graphicx}
\usepackage{xcolor}
\usepackage{colortbl}
\usepackage{enumitem}
\usepackage[most]{tcolorbox}

\definecolor{promptblue}{RGB}{80, 100, 200}
\newtcolorbox{promptbox}[2][]{
    enhanced,
    colback=white,                 
    colframe=black!75,             
    boxrule=0.8pt,                 
    arc=3mm,                       
    auto outer arc,                
    fonttitle=\bfseries,           
    colbacktitle=promptblue,       
    coltitle=white,                
    breakable,                     
    title={#2},                 
    #1                             
}
\def\paperID{14569} 
\def\confName{CVPR}
\def\confYear{2026}

\title{From Frames to Clips: Training-free Adaptive Key Clip Selection for Long-Form Video Understanding}

\author{
Guangyu Sun\textsuperscript{1,2}\thanks{Work done during an internship at Amazon Prime Video.},
Archit Singhal\textsuperscript{1},
Burak Uzkent\textsuperscript{1}, 
Mubarak Shah\textsuperscript{1,2}, 
Chen Chen\textsuperscript{2}, 
Garin Kessler\textsuperscript{1}
\\\\
\textsuperscript{1} Amazon \quad
\textsuperscript{2} University of Central Florida \\
\small{\tt{\{sgy, architsl, burauzke, kesslerg\}@amazon.com; \{shah, chen.chen\}@crcv.ucf.edu}} 
}

\begin{document}
\maketitle

\begin{abstract}
Video Large Language Models (VLMs) have achieved strong performance on various vision-language tasks, yet their practical use is limited by the massive number of visual tokens produced from raw video frames, which quickly exhausts the model’s context window. Existing solutions mitigate this issue by selecting a sparse set of frames, but such frame-wise selection discards essential temporal dynamics in long-form videos, leading to suboptimal reasoning about motion and event continuity. In this work, we systematically examine the role of temporal information and show that extending selection from isolated key frames to temporally coherent key clips improves video understanding. To maintain a fixed computational budget while accommodating the larger token footprint of clips, we introduce frame resolution as a controllable factor in frame selection, enabling a trade-off between spatial resolution and clip length. Building on this idea, we propose an adaptive clip length module that dynamically balances these factors to ensure a constant token count per video. Experiments on three long-form video benchmarks demonstrate that our training-free approach, F2C, outperforms uniform sampling by up to $8.1\%$, $5.6\%$, and $10.3\%$ on Video-MME, LongVideoBench, and MLVU, respectively. These results highlight the importance of preserving temporal coherence in frame selection and provide a practical pathway for scaling VLMs to real-world video understanding applications.
\end{abstract}

\section{Introduction}
\label{sec:intro}

Video Large Language Models (VLMs) have demonstrated strong capabilities in video understanding tasks such as video question answering (VQA) and captioning~\citep{maaz2023video,lin2023mm,zhang2023simple,lin2023video,jin2024chat,ren2024timechat,Liu_2024_CVPR}. By integrating visual encoders with Large Language Models (LLMs), VLMs can reason over multimodal content and achieve impressive performance~\citep{yangVid2SeqLargeScalePretraining2023, chenShareGPT4VideoImprovingVideo2024, wuDIBSEnhancingDense2024, kimImageGridCan2024, minMoReVQAExploringModular2024}. However, processing long-form videos remains a major bottleneck due to the overwhelming number of visual tokens relative to language tokens. For instance, a video with 1K resolution (1024$\times$768) produces 4,015 tokens per two frames when processed by Qwen2.5-VL~\citep{bai2025qwen25vltechnicalreport}. Even when sampled at only 1 FPS, a one-hour video yields over 7 million tokens, far exceeding the context length limits of current LLMs. Beyond computational constraints~\citep{yuFrameVoyagerLearningQuery2024,shenLongVUSpatiotemporalAdaptive2024}, excessive visual input can distract model attention, making it difficult to identify relevant content when facing more than 2k tokens~\citep{vilamp}. 

\begin{figure}[t]
\centering
    \includegraphics[width=\linewidth]{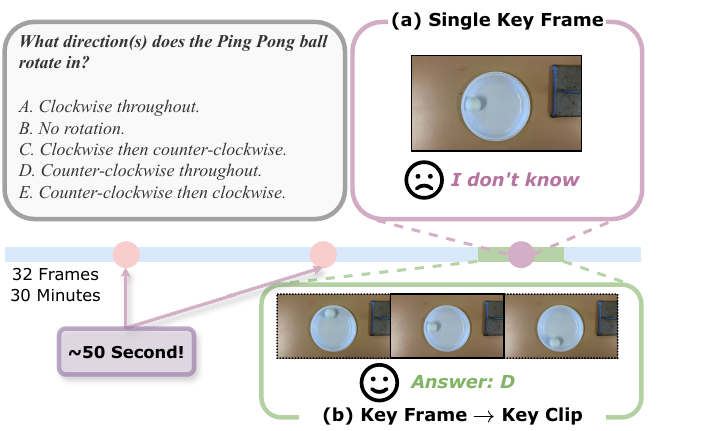}
    \vspace{-5mm}
    \caption{\textbf{Motivation of key clip selection.} In long-form videos (\eg, 30 minutes), uniform sampling into 32 frames yields gaps of more than 50 seconds, so only a single frame may be selected from an event. As an example from~\cite{shangguan2025tomatoassessingvisualtemporal} shown, relying on one key frame (a) makes it impossible to answer motion-related questions such as the rotation direction of a Ping Pong ball. By extending a key frame into a short key clip (b), temporal continuity is preserved, enabling the correct reasoning. 
    }
    \label{fig:motivation}
\end{figure}

To mitigate this, most VLMs adopt \textit{uniform sampling} to reduce the number of frames before encoding. While effective for short video clips, uniform sampling assumes equal importance across all frames, which is rarely the case in long and uncurated videos such as surveillance or instructional recordings. In such scenarios, uniformly sampled frames often capture irrelevant content, degrading downstream performance~\citep{li2023videochat,shuVideoXLExtraLongVision2024,BOLT}.

Recent works have explored keyframe selection as a form of \textit{context management} for long videos~\citep{ataallahGoldfishVisionLanguageUnderstanding2024a, yuFrameVoyagerLearningQuery2024, shenLongVUSpatiotemporalAdaptive2024}. These methods typically rank frames based on visual similarity to the query and select the most relevant ones. Although this improves efficiency, the approaches overlook temporal continuity. In long videos, sparse keyframe sampling can leave large temporal gaps. For instance, 32 frames sampled from a 30-minute video correspond to roughly 50 seconds between frames, which loses motion cues and event progression, as shown in Figure~\ref{fig:motivation}. This limitation raises a natural question: \textit{Can temporal information from frames adjacent to the key frames improve downstream performance in VLMs?}




In this work, we revisit the frame selection problem from a new perspective. Instead of relying solely on isolated frames, we propose selecting \textit{key clips}, which are short temporal segments centered around key frames. This preserves both semantic relevance and local temporal continuity.  
However, naively adding more frames increases the number of visual tokens, intensifying the memory bottleneck. A practical solution to this problem is to compensate frames with \textit{lower spatial resolutions} for longer clip length. This introduces resolution as an additional factor in frame selection. By lowering resolution, one can incorporate longer clips without increasing the total token count. Thus, a principled trade-off emerges between clip length and spatial resolution, allowing a spatio-temporal balance.

Motivated by these insights, we propose \textbf{Frames-to-Clips} (F2C), a training-free framework
for key clip selection with adaptive resolution to avoid additional computation. F2C enhances temporal coherence while maintaining
computational efficiency, offering a practical solution to scaling VLMs for long-form video understanding.
Through empirical analysis on long-form VQA benchmarks, we compare F2C with other frame selection methods and find that the temporal continuity in key clips consistently improves performance, underscoring the importance of temporal context in long-form video understanding.

Our contributions are three-fold:
\begin{itemize}[leftmargin=*, topsep=0pt, itemsep=1pt, partopsep=0pt, parsep=0pt]
    \item We provide a comprehensive analysis of frame selection strategies and show that selecting \emph{temporally coherent key clips}, rather than isolated frames, substantially improves VLM performance on long-form videos.
    \item To manage the increased tokens introduced by clips, we propose F2C, a key clip selection framework enabling an adaptive trade-off between spatial resolution and temporal length for each clip.
    \item \emph{Without training}, F2C consistently outperforms uniform sampling, improving performance by up to 8.1\%, 5.6\%, and 10.3\% on Video-MME~\citep{fu2024video}, LongVideoBench~\citep{wu2024longvideobench}, and MLVU~\citep{zhou2024mlvu}, respectively, offering an efficient and scalable solution for long-form video understanding.
    
\end{itemize}

\begin{figure*}[!t]
\centering
\begin{minipage}{0.48\textwidth}
    \centering
    \includegraphics[width=0.85\linewidth]{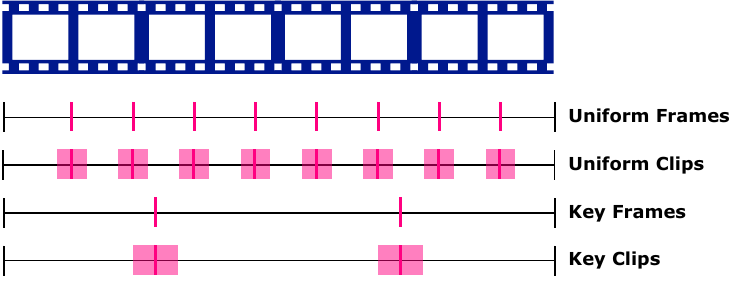}
    \vspace{-3mm}
    \caption{\small Uniform Sampling samples frames evenly from the video, and Key Frames are human-annotated crucial frames to answer the question. Uniform and Key Clips augment frames with their neighboring frames.}
    \label{fig:methods}
\end{minipage}
\hfill
\begin{minipage}{0.48\textwidth}
    \begin{table}[H]
    \centering
    \vspace{-5mm}
    \caption{Comparison of different sampling methods on the Ego4D-Haystack dataset using Qwen2.5-VL-7B.}
\label{tab:haystack}
\vspace{2mm}
    \resizebox{0.75\linewidth}{!}{%
        \begin{tabular}{lcc}
        \toprule
        \textbf{Method} & \textbf{\# Frames} & \textbf{QA Accuracy \%} \\
        \midrule
        Uniform   & 32 & 45.2 \\
        Uniform   & 64 & 49.8 \\
        Uniform Clips & 32 $\times$ 2 & 50.4\\ 
        \midrule
        Key Frames  & 3-4 & 63.0 \\
        Key Clips   & 32 & 65.1 \\
        \bottomrule
        \end{tabular}
    }
    \end{table}
\end{minipage}%
\vspace{-5mm}
\end{figure*}

\section{Related Work}

\subsection{VLM for Long-form Video}

Recent advances in VLMs have extended the multimodal reasoning ability of LLMs to video understanding tasks such as question answering and captioning~\citep{song2025moviechat}. Many works improve the backbone architecture to better handle long-form inputs: Video-LLaVA~\citep{lin2023video}, Qwen2.5-VL~\citep{bai2025qwen25vltechnicalreport}, and VideoChat2~\citep{li2023videochat} adapt image-language models with temporal modules or tailored training strategies; Chat-UniVi~\citep{jin2024chat} and VideoLLaMA2~\citep{chengVideoLLaMA2Advancing2024} refine multimodal fusion and instruction tuning; LLaVA-NeXT-QW2~\citep{liu2024llavanext}, LLaVA-OneVision~\citep{liLLaVAOneVisionEasyVisual2024a}, and SlowFast-LLaVA~\citep{xu2024slowfast,xu2025slowfastllava15familytokenefficientvideo} and $\infty$-Video~\cite{dos2025video} enhance vision encoders or incorporate multi-scale temporal modeling; and LongVILA~\citep{xueLongVILAScalingLongContext2024}, LongVA~\citep{zhang2024longva}, Video-XL~\citep{shuVideoXLExtraLongVision2024}, Video-CCAM~\citep{fei2024videoccamenhancingvideolanguageunderstanding}, LongVU~\citep{shenLongVUSpatiotemporalAdaptive2024}, ViLAMP~\citep{vilamp}, and Keye-VL~\citep{keye} explicitly target long videos through hierarchical attention, temporal compression, or improved long-context modeling. Despite these advances, most approaches still rely on \textit{uniform sampling} to reduce raw videos into a fixed number of frames, overlooking redundancy and the uneven distribution of relevant information. In contrast, we focus on this preprocessing stage and propose a context management strategy that selects temporally coherent clips with adaptive resolution, providing more informative and efficient inputs without modifying the model architecture.

\subsection{Training-based Frame Selection}

Learning-based approaches have been proposed to select the most informative frames for VLMs, though they are typically limited to short videos due to computational cost. ATP~\citep{buch2022revisitingvideovideolanguageunderstanding} and FFS~\citep{buch2025flexible} train selectors end-to-end using downstream task losses, while Frame-Voyager~\citep{yuFrameVoyagerLearningQuery2024} ranks frame candidates by their task loss. \cite{Hu2025} leverage a strong vision-language model to generate pseudo-labels for supervision. ViaRL~\citep{xu2025viarladaptivetemporalgrounding}, ReFoCUS~\citep{lee2025refocusreinforcementguidedframeoptimization} and SeViLA~\citep{yu2023self} apply reinforcement learning and self-learning to improve grounding before answering. GenS~\citep{yao2025generative} and Chain-of-Frames~\citep{ghazanfari2025chainofframesadvancingvideounderstanding} introduce datasets with keyframe annotations to enable supervised training. Although effective in controlled settings, these methods are resource-demanding to train and scale poorly to long-form videos, making training-free selection strategies more practical.

\subsection{Training-free Frame Selection}

Several methods have explored training-free frame selection. 
BOLT~\citep{BOLT} is an approach that leverages inverse transform sampling to improve diversity in selected frames. 
TCoT~\citep{arnab2025temporal} and CoS~\citep{hu2025coschainofshotpromptinglong} later utilized the long context window of models like Gemini-1.5-Flash~\citep{geminiteam2024gemini15unlockingmultimodal} to reason over candidate keyframes before answering. 
MDP3~\citep{sun2025mdp3trainingfreeapproachlistwise} advanced this direction with a list-wise selection strategy that balances query relevance, diversity, and sequentiality. 
AKS~\citep{aks} formulates keyframe selection as an optimization over prompt relevance and video coverage, and proposes an adaptive algorithm that selects informative keyframes to maximize retained information under fixed token budgets. 
Most recently, Q-Frame~\citep{zhang2025qframequeryawareframeselection}, which is the most related to our work, incorporated dynamic resolution by ranking frames into multiple resolution levels. 
In contrast, we rethink frame selection from a dynamic perspective and propose an adaptive strategy over key clip length rather than treating frames as isolated units.

\section{Preliminary}
\subsection{Problem Formulation}
In this work, we consider a Video Question Answering (VQA) task for a video $V$ represented with $N$ frames $V=\{f_1,f_2,\dots,f_N\}$
 and an associated text query $Q$.
 VQA aims to generate an answer $A$ that accurately addresses the query based on the content in the given video. Our goal is that given $(V, Q)$, we want to sample critical frames to generate the answer with accurate information.

Due to the limitations of the computation and memory, the VLM solutions for VQA usually down-sample the video into $K$ frames based on the hardware constraints. $K$ will usually be 8 to 32 frames, \ie, $K\ll N$.  By default, the frames will be sampled \textit{uniformly}. Selected frames are then passed to a VLM along with the text query $Q$ to get an answer. Formally, a VLM represented by $f_{\text{VLM}}$ generates an answer for the given input as
\begin{equation}
A = f_{\text{VLM}}(\emph{Select}(V), Q).
\end{equation}
In this work, 
our goal is to perform input selection, $\emph{Select}$, to provide more useful information to the VLM.
Specifically, our task is formulated as choosing a set of selected frames $\{f'_1,f'_2,\dots,f'_K\}$, where ${f'}_i \in \mathbb{R}^{H\times W}$, together with their spatial resolution represented by $H$ and $W$. 

So far, we have only selected frames with their associated spatial resolution. However, temporal information can be crucial to answer certain questions (see Figure~\ref{fig:motivation}). We can incorporate temporal information, \emph{key clips}, into our input for VLM as $\{C_1,C_2,\dots,C_K\}$, where ${C}_i \in \mathbb{R}^{H\times W\times L}$. In this study, we propose an adaptive approach to incorporate \emph{key clips} with adaptive spatial ($H, W$) and temporal resolution ($L$).


\begin{figure*}[t]
\centering
    \includegraphics[width=\linewidth]{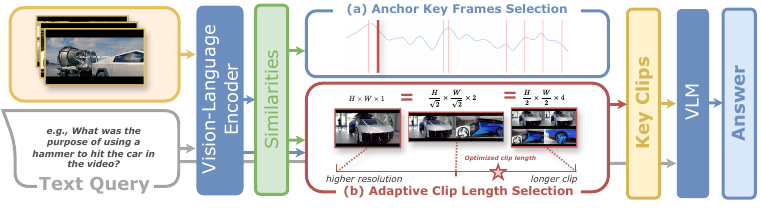}
\caption{\textbf{Overview of F2C.} A video and text query are first processed by a vision-language encoder to select (a) \textit{anchor key frames}. Each anchor is then extended into a short \textit{key clip} through (b) adaptive clip length selection, which adjusts the trade-off between resolution and clip length under the same budget. The constructed key clips preserve temporal continuity and, together with the query, are fed into a VLM to generate the answer.}
    \label{fig:overview}
\end{figure*}

\subsection{Key Frames and Temporal Information}

To investigate the impact of key frames and the temporal information in VQA, we conduct an empirical study on the Ego4D-Haystack~\citep{tstar} dataset, which provides \textit{human-annotated ground-truth} key frames for each video containing sufficient information to answer the question. This dataset contains long-form videos, each approximately 30 minutes in duration, which requires demanding capability for long-context understanding. We use Qwen2.5-VL-7B as the downstream VLM to evaluate performance on the VQA task under uniform sampling, ground-truth key frames, and augmented key clips, as shown in Figure~\ref{fig:methods}.

As shown in Table~\ref{tab:haystack}, selecting key frames based on human annotations yields a substantial improvement over uniform frame sampling, underscoring the importance of constructing task-relevant visual context. Note that even with only 3-4 key frames, key frame selection can outperform uniform sampling even with 64 frames, highlighting the strong effectiveness of selecting related frames.

To further investigate the role of temporal information, we extend both uniform sampling and keyframe selection by including neighboring frames to form short, temporally coherent segments, denoted as \emph{uniform clips} and \emph{key clips}. For uniform sampling, each frame is augmented into a 2-frame clip, while for keyframe selection, we expand to a total of 32 frames for fair comparison. This simple augmentation consistently improves VQA accuracy, showing that preserving local temporal continuity benefits multimodal reasoning for both uniform sampling and key frames.

These results motivate our proposed method, which generalizes this idea by selecting key clips, thereby capturing both semantic relevance and temporal structure.

\section{Proposed Method}

Since temporal information can better help the QA task, we propose F2C, from key frame selection to a perspective of key clip selection. We decompose the key clips selection into two subtasks: 1)~Select $K_\text{anchor}$ initial key frames that become the center of the key clips (\textit{anchor key frames}), $\{C_1,C_2,\dots,C_{K_\text{anchor}}\}$, and 2)~Determine the clip length, $L_i$, for each anchor key frame to construct the key clip, as shown in Figure~\ref{fig:overview}.


\subsection{Anchor Key Frame Selection}

To construct key clips, the first step is to determine the anchor key frames that serve as the centers of the clips. To control the diversity of the selected frames, we define the number of anchor key frames as $K_\text{anchor}$.
We follow two guiding principles: (1)~\textbf{Relevancy}, filtering out frames that are not related to the query; and (2)~\textbf{Diversity}, ensuring that the selected anchors cover different parts of the video to avoid redundancy, especially since temporal neighbors will later be added to each anchor.

For relevancy, we leverage a pre-trained contrastive vision-language encoder $E$ (\eg, CLIP~\citep{radford2021learning} or SigLIP~\citep{zhai2023sigmoid,tschannen2025siglip2multilingualvisionlanguage}) to compute the cosine similarity between each frame $f_i$ and the text query $Q$ as
\begin{equation}
\label{eq:rel}
    r(f_i) = \cos(E(f_i), E(Q)).
\end{equation}
For multiple-choice questions, both the question and the candidate options are included in the query to avoid missing crucial clues.

Once relevancy scores are computed, anchor frames are selected by combining relevance with diversity. Inspired by the Watershed algorithm~\citep{haralick1987image}, we first identify local maxima of the similarity curve (\ie, watersheds) as an initial candidate pool. To further ensure diversity, we apply K-means clustering on the temporal indices of these candidates into $K_\text{anchor}$ clusters, and select within each cluster the frame with the highest similarity score as the anchor key frame. We provide a pseudo-code of the algorithm in the \textit{supplementary}~\ref{app:water}.

\subsection{Adaptive Clip Length Selection}
\subsubsection{Trade-off between Resolution and Clip Length}

After selecting the anchor key frames, we analyze the trade-off between frame resolution and clip length under a fixed token budget. 
Consider a selection of $K$ frames of ${f}_i \in \mathbb{R}^{H\times W}$.
In this case, our budget, $B$, can be formulated as
\begin{equation}
\textstyle
    B = \frac{K\cdot  H \cdot W}{Z},
\end{equation}
where $Z$ is the number of pixels per token. As seen in the equation, the total number of visual tokens is proportional to both $K$ and their spatial resolution, $H$ and $W$.

Now consider constructing $K_{\text{anchor}}$ anchor frames, each downsampled by a factor of $s$ to resolution $(H/s, W/s)$, and extending each anchor into a key clip of length $L$. 
The resulting token budget becomes
\begin{equation}
\textstyle
    B_{\text{clip}} = 
    \frac{K_{\text{anchor}}\cdot L\cdot (H/s)\cdot (W/s)}{Z}
    = \frac{K_{\text{anchor}} \cdot L}{s^2 \cdot K} \cdot B.
\end{equation}

To keep the overall token budget unchanged (\ie, $B_{\text{clip}}=B$), the clip length must satisfy
\begin{equation}
\label{eq:trade-off}
\textstyle
    L = \frac{s^2 \cdot K}{K_{\text{anchor}}}.
\end{equation}

This relation reveals a fundamental trade-off: longer clips can only be obtained by either reducing spatial resolution or decreasing the number of anchors, while higher-resolution clips limit the temporal context that can be included. 
This observation motivates our adaptive resolution strategy, which dynamically balances resolution and clip length under a fixed budget.

\begin{table*}[t]
\centering
\small
\caption{\small \textbf{Evaluation results of VQA accuracy across different frame budgets on Video-MME, LongVideoBench (LVB), and MLVU.} Missing results are shown as ``-''. $^\dagger$ means our re-implementation results. * means \# frames as a hyperparameter. Methods with $^\ddagger$ uses various resolutions, and their budgets are equivalent to the reported \# frames with the original resolution.}\label{tab:main}
\setlength{\tabcolsep}{8pt}
\resizebox{0.95\linewidth}{!}{%
\begin{tabular}{lcccccccc}
\toprule
\multirow{2}{*}{\textbf{Method}} 
  & \multirow{3}{*}{\textbf{Size}} 
  & \multirow{3}{*}{\textbf{\# Frames}} 
  & \multicolumn{4}{c}{\textbf{Video-MME (w.o./w. sub.)}} 
  & \multirow{2}{*}{\textbf{LVB}} 
  & \multirow{2}{*}{\textbf{MLVU}} \\
 & & & Short & Medium & Long & Overall & & \\
  \textit{Avg. Video Duration}& & & \textit{1.3min} & \textit{9min} & \textit{41min} & \textit{17min} & \textit{12min} & \textit{12min}\\
\midrule
\rowcolor{gray!30}\multicolumn{3}{l}{\textit{Training-based VLM w/ Uniform Sampling:}}&&&&&&\\
Video-LLaVA~\citep{lin2023video} & 7B & 8 & 45.3 / 46.1 & 38.0 / 40.7 & 36.2 / 38.1 & 39.9 / 41.6 & 39.1 & 47.3 \\
Qwen-VL~\citep{baiQwenVLVersatileVisionLanguage2023} & 7B& 8 & 46.9 / 47.3 & 38.7 / 40.4 & 37.8 / 37.9 & 41.1 / 41.9 & - & - \\
VideoChat2~\citep{li2023videochat} & 7B& 8 & 48.3 / 52.8 & 37.0 / 39.4 & 33.2 / 39.2 & 39.5 / 43.8 & 39.3 & 44.5 \\
Chat-UniVi-V1.5~\citep{jin2024chat} & 7B& 8 & 45.7 / 51.2 & 40.3 / 44.6 & 35.8 / 41.8 & 40.6 / 45.9 & - & - \\
VideoLLaMA2~\citep{chengVideoLLaMA2Advancing2024} & 7B& 16 & 56.0 / - & 45.4 / - & 42.1 / - & 47.9 / - & - & - \\
LLaVA-NeXT-QW2~\citep{liu2024llavanext} & 7B& 8 & 58.0 / - & 47.0 / - & 43.4 / - & 49.5 / - & - & - \\
LongVILA~\citep{xueLongVILAScalingLongContext2024} & 8B& 128 & 60.2 / - & 48.2 / - & 38.8 / - & 49.2 / - & - & - \\
LongVA~\citep{zhang2024longva} & 7B & 128& 61.1 / 61.6 & 50.4 / 53.6 & 46.2 / 47.6 & 52.6 / 54.3 & - & - \\
Video-XL~\citep{shuVideoXLExtraLongVision2024}& 7B & 128/256 & 64.0 / 67.4 & 53.2 / 60.7 & 49.2 / 54.9 & 55.5 / 61.0 & - & 64.9 \\
LLaVA-OneVision~\citep{liLLaVAOneVisionEasyVisual2024a} & 7B & * & 64.0 / 67.4 & 53.2 / 60.7 & 49.2 / 54.9 & 58.2 / - & 56.3 & 64.7 \\
Video-CCAM~\citep{fei2024videoccamenhancingvideolanguageunderstanding} & 9B& 96 & 61.9 / 63.1 & 49.2 / 52.3 & 39.6 / 42.4 & 50.3 / 52.6 & - & 58.5 \\
LongVU~\citep{shenLongVUSpatiotemporalAdaptive2024} & 7B & 1fps& 64.7 / - & 58.2 / - & 59.5 / - & 60.9 / - & - & 65.4 \\
SF-LLaVA-1.5~\citep{xu2025slowfastllava15familytokenefficientvideo} & 7B & 128& - / - & - / - & - / - & 63.9 / - & 62.5 & 71.5\\
ViLAMP~\citep{vilamp} & 7B & 1fps & - / - & -/ - & 57.8 / 68.1 & 67.5 / 73.5 & 61.2 & -\\
Keye-VL-1.5~\citep{keye} & 8B & 64& 81.2	/ 84.0&	70.7 / 71.9 & 67.1 / 66.2 & 73.0 / 74.0 & 66.0 & -\\

\midrule

\rowcolor{gray!30}\multicolumn{3}{l}{\textit{Training-based Frame Selection:}}&&&&&&\\
Frame-Voyager~\citep{yuFrameVoyagerLearningQuery2024} & 7B & 8& 67.3 / -  & 56.3 / -  & 48.9 / -  & 57.5 / -  & - & 65.6 \\
Hu \etal~\cite{Hu2025} & 8.5B& 128 &  69.6 / -  & 54.1 / -  & 51.9 / -  & 58.7 / -  & - & -\\
GenS~\citep{yao2025generative}& 7B & 54& - / - & - / -  & - / -  & - / - & 58.7 & 64.8\\
\midrule
\rowcolor{gray!30}\multicolumn{3}{l}{\textit{Training-free Frame Selection:}}&&&&&&\\
\multicolumn{3}{l}{\textbf{Qwen2.5-VL – 8 Frames}} \\
+ \textit{Uniform}    & 7B& 8 & 60.9 / 68.6 & 51.4 / 56.2 & 47.4 / 51.2 & 53.3 / 58.7 & 53.3 & 52.8 \\
+ \textit{Top-k }     & 7B& 8 & 66.8 / 71.0 & 55.6 / 58.0 & 47.7 / 51.0 & 56.7 / 60.0 & 58.7 & 61.1 \\
+ \textit{Watershed}  & 7B& 8 & 67.2 / 69.4    & 54.8 / 58.6    & 48.0 / 51.8    & 56.7 / 59.9    & 56.8 & 60.0 \\
+ \textit{BOLT $^\dagger$}~\citep{BOLT}       & 7B& 8 & 66.0 / 71.7 & 54.6 / 57.1 & 50.4 / 52.0 & 57.0 / 60.3 & 55.6 & 59.0 \\
+ \textit{AKS$^\dagger$}~\citep{aks} & 7B & 8 & 62.1 / 66.2 & 51.4 / 55.4 & 48.8 / 49.8 & 54.1 / 57.1 & 52.0 &53.7 \\
+ \textit{Q-Frame$^\dagger$}~\citep{zhang2025qframequeryawareframeselection}& 7B & 8$^\ddagger$ & 71.9 / 74.0 & 58.2 / 60.6& 50.7/ 54.0 & 60.3 / 62.9 & 56.6&58.2\\
+ \textit{Ours}       & 7B& 8$^\ddagger$ & \textbf{72.4} / \textbf{74.1} & \textbf{60.7} / \textbf{63.1} & \textbf{51.2} / \textbf{55.0} & \textbf{61.4} / \textbf{64.1} & \textbf{58.9} & \textbf{63.1} \\
\midrule
\multicolumn{3}{l}{\textbf{Qwen2.5-VL – 16 Frames}}  \\
+ \textit{Uniform}    & 7B& 16 & 67.3 / 71.2 & 55.0 / 58.8 & 48.9 / 51.9 & 57.1 / 60.6 & 56.7 & 57.1 \\
+ \textit{Top-k}      & 7B& 16 & 70.2 / 72.0 & 57.4 / 59.1 & 50.8 / 51.7 & 59.5 / 60.9 & 59.1 & 63.9 \\
+ \textit{Watershed}  & 7B& 16 & 70.0 / 72.1 & 58.3 / 62.7 & 50.7 / 52.4 & 59.7 / 62.4 & 56.8 & 62.3 \\
+ \textit{BOLT $^\dagger$}~\citep{BOLT}       & 7B& 16 & 71.7 / 72.3 & 57.7 / 60.8 & 49.7 / 53.1 & 59.7 / 62.1 & 58.0 & 64.5 \\
+ \textit{AKS$^\dagger$}~\citep{aks} & 7B & 16 
& 66.4 / 70.2 
& 56.9 / 60.2 
& 48.6 / 53.1 
& 57.3 / 61.2
& 55.7
& 57.8\\
+ \textit{Q-Frame$^\dagger$}~\citep{zhang2025qframequeryawareframeselection}& 7B & 16$^\ddagger$ & 73.4 / 74.9 & 61.4 / 63.3 & 52.3 / 53.7 & 62.4 / 64.1 & 57.1 & 61.9\\
+ \textit{Ours}       & 7B& 16$^\ddagger$ & \textbf{73.8} / \textbf{75.6} & \textbf{62.2} / \textbf{65.6} & \textbf{54.1} / \textbf{56.1} & \textbf{63.4} / \textbf{65.7} & \textbf{61.1} & \textbf{65.5} \\
\midrule
\multicolumn{3}{l}{\textbf{Qwen2.5-VL – 32 Frames}}  \\
+ \textit{Uniform}     & 7B& 32 & 72.6 / 73.7 & 59.0 / 62.6 & 51.8 / 55.1 & 61.1 / 63.8 & 58.4 & 59.4 \\
+ \textit{Top-k }      & 7B& 32 & 74.2 / \textbf{76.4} & 59.9 / 62.0 & 51.9 / 54.0 & 62.0 / 64.1 & 60.1 & 66.6 \\
+ \textit{Watershed}   & 7B& 32 & 71.8 / 73.6 & 60.2 / 63.4 & 52.1 / 55.2 & 61.4 / 64.1 & 58.5 & 64.0 \\
+ \textit{BOLT $^\dagger$}~\citep{BOLT}        & 7B& 32 & \textbf{74.3} / 76.2 & 64.2 / 63.9 & 53.8 / 56.4 & 64.1 / 65.5 & 58.6 & 66.3 \\
+ \textit{AKS$^\dagger$}~\citep{aks} & 7B & 32
& 66.4 / 76.0
& 56.9 / 65.0
& 48.6 / 55.4
& 57.3 / 65.5 
& 58.3
&63.1\\
+ \textit{Q-Frame$^\dagger$}~\citep{zhang2025qframequeryawareframeselection} & 7B & 32$^\ddagger$ & 73.0 / 75.0 & 61.7 / 62.9 & 53.1 / 53.7 & 62.6 / 63.9 & 58.7 & 41.6 \\
+ \textit{Ours}        & 7B& 32$^\ddagger$ & 73.8 / 75.9 & \textbf{66.3} / \textbf{68.1} & \textbf{56.6} / \textbf{57.8} & \textbf{65.6} / \textbf{67.3} & \textbf{60.8} & \textbf{66.8} \\
\bottomrule

\end{tabular}
}
\end{table*}

\begin{figure*}[t]
\centering
\begin{minipage}{0.48\linewidth}
    \centering
    \includegraphics[width=\linewidth]{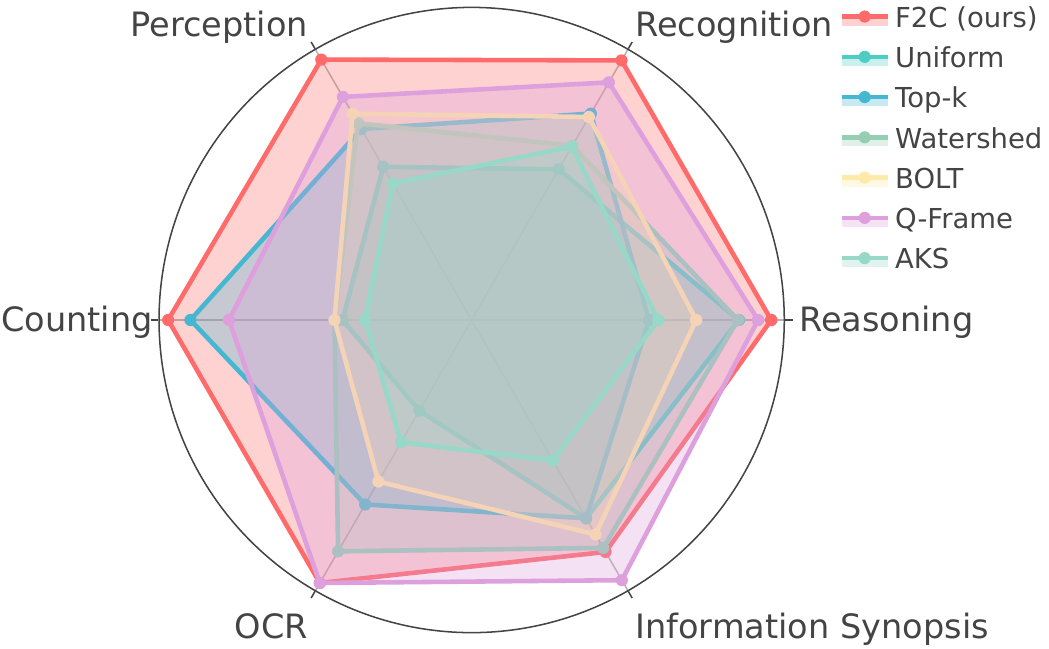}
    \vspace{-2mm}
    \caption{Performance on each type of questions on Video-MME with 16 frames.}
    \label{fig:radar}
\end{minipage}
\hfill
\begin{minipage}{0.48\linewidth}

\begin{minipage}{\linewidth}
\vspace{-5mm}

\begin{table}[H]
    \centering
    \small
        \vspace{-3mm}
    \caption{Ablation study on Video-MME.}
\label{tab:ablation}
\resizebox{\linewidth}{!}{%
\begin{tabular}{lcccc}
    \toprule
    \textbf{Method} & \textbf{Short} & \textbf{Medium} & \textbf{Long} & \textbf{Overall} \\
    \midrule
    Uniform & 67.3  & 55.0  & 48.9  & 57.1  \\
    Watershed & 70.0  & 58.3  & 50.7  & 59.7 \\
    + Key Clip (Fixed) & 73.0 & 60.0 & 53.7 & 62.2 \\
    + Key Clip (Adaptive) & \textbf{73.8} & \textbf{62.2} & \textbf{54.1} & \textbf{63.4} \\
    \bottomrule
    \end{tabular}
}
\end{table}
\end{minipage}

\begin{minipage}{\linewidth}
\begin{table}[H]
    \centering
        \caption{\small \textbf{Impact of temporal continuity.} We evaluate Baselines use 64 isolated low-resolution frames, while F2C uses 16 clips of up to 4 frames. The subscript means the performance delta compared to using 16 high-resolution frames.}
    \label{tab:temporal}
    \resizebox{\linewidth}{!}{%
\begin{tabular}{lccccc}
\toprule
\textbf{Method} &\textbf{\# Frames} & \textbf{Short} & \textbf{Medium} & \textbf{Long} & \textbf{Overall} \\
\midrule
Uniform & 64 
& 71.3\textsubscript{\textcolor{ForestGreen}{+4.0}} 
& 60.1\textsubscript{\textcolor{ForestGreen}{+5.1}} 
& 51.4\textsubscript{\textcolor{ForestGreen}{+2.5}} 
& 61.0\textsubscript{\textcolor{ForestGreen}{+3.9}}  \\

Top-$k$ & 64 
& 73.6\textsubscript{\textcolor{ForestGreen}{+3.4}} 
& 59.9\textsubscript{\textcolor{ForestGreen}{+2.5}} 
& 52.3\textsubscript{\textcolor{ForestGreen}{+1.5}} 
& 61.9\textsubscript{\textcolor{ForestGreen}{+2.4}}  \\

BOLT & 64
& 73.4\textsubscript{\textcolor{ForestGreen}{+1.7}} 
& 61.0\textsubscript{\textcolor{ForestGreen}{+3.3}} 
& 50.6\textsubscript{\textcolor{ForestGreen}{+0.9}} 
& 61.7\textsubscript{\textcolor{ForestGreen}{+2.0}}  \\

Watershed & 64
& 72.3\textsubscript{\textcolor{ForestGreen}{+2.3}} 
& 59.8\textsubscript{\textcolor{ForestGreen}{+1.5}} 
& 53.4\textsubscript{\textcolor{ForestGreen}{+2.7}} 
& 61.9\textsubscript{\textcolor{ForestGreen}{+2.2}}  \\

\textbf{Ours} & $\le$16$\times$4 
& \textbf{73.8} 
& \textbf{62.2} 
& \textbf{54.1} 
& \textbf{63.4} \\
\bottomrule
\end{tabular}
    }
\end{table}
\end{minipage}
\end{minipage}
\vspace{-5mm}
\end{figure*}





\subsubsection{Clip-Specific Selection}

In the previous subsection, we applied a fixed clip length $L$ for all clips, which treats them equally and ignores their distinct characteristics. To address this, we extend $L$ into a clip-specific length $l_i$ for each clip $C_i$, centered at anchor frame $k_i$. Similar to Equation~\ref{eq:trade-off}, $l_i$ is constrained by its scaling factor $s_i$ as
$
    l_i = \frac{s_i^2 K}{K_\text{anchor}}.
$

To reduce the search space, we introduce a maximum scaling factor $s_\text{max}$ as a hyperparameter, which defines the largest possible clip length 
$
    l_\text{max} = \frac{s_\text{max}^2 K}{K_\text{anchor}}.
$
Each clip is then optimized independently within the range $[1, l_\text{max}]$.

For a candidate clip length $l$, the clip $C_i$ spans the interval $[k_i-\lfloor(l-1)/2\rfloor, k_i+\lfloor(l-1)/2\rfloor]$. Its importance is measured by three components:

\noindent\textbf{Relevancy.} We compute the average cosine similarity between the frames in $C_i$ and the text query:
\begin{equation}
\textstyle
S_C(l) = \frac{1}{l}\sum_i r(f_i),
\end{equation}
where $r(f_i)$ is defined in Equation~\ref{eq:rel}.

\noindent\textbf{Redundancy.} To discourage redundant frames, we compute the average pairwise similarity between frames:
\begin{equation}
\textstyle
R_C(l) = \frac{1}{l(l-1)}\sum_i \sum_{j \ne i} \cos\big(E(f_i), E(f_j)\big),
\end{equation}
where $E$ is the vision-language encoder.

\noindent\textbf{Temporal reward.} We encourage longer clips to compensate for potential noise in similarity scores by adding a reward proportional to the relative clip length $l/l_\text{max}$.

\noindent\textbf{Optimization objective.} Combining the above, the clip is determined by finding $l^{*}$ that optimizes the following by \textit{exhaustive} search
\begin{equation}
\textstyle
    l^* = \arg\max_{1 \le l \le l_\text{max}} \Big(S_C(l) - \lambda_r R_C(l) + \lambda_l \tfrac{l}{l_\text{max}}\Big),
\end{equation}
where $\lambda_r$ and $\lambda_l$ control the scales between each score. The corresponding scaling factor is then computed as
$
    s^*_i = \sqrt{\tfrac{K_\text{anchor} \cdot l^*_i}{K}}
$ to obtain the spatial resolution of the current clip.

Finally, overlapping clips with identical resolutions are merged to avoid redundancy, yielding the set of key clips that are passed into the VLM.

\section{Experiments}

\begin{figure*}[t]
\centering
\small
\begin{minipage}{0.48\linewidth}
    \centering
    \includegraphics[width=0.85\linewidth]{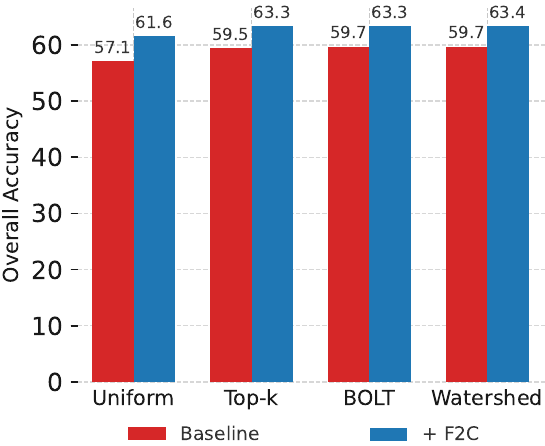}
    \vspace{-3mm}
    \caption{Impact of different initial keyframe selectors on Video-MME with $K=16$ and extending them to key clips by F2C.}
    \label{fig:initial}
\end{minipage}
\hfill
\begin{minipage}{0.48\linewidth}
    \centering
        \begin{table}[H]
        \setlength{\tabcolsep}{2pt}
    \centering
    \vspace{-5mm}
        \caption{Token number analysis on Video-MME under different budgets of full-resolution frames.}
    \label{tab:token_efficiency}
        \resizebox{\linewidth}{!}{%

    \begin{tabular}{lccc}
    \toprule
    \textbf{Method} & \textbf{8 Frames} & \textbf{16 Frames} & \textbf{32 Frames} \\
    \midrule
    Baselines & 4428.97 & 8276.95 & 17259.13 \\
    Ours & 4359.95\textsubscript{\textcolor{ForestGreen}{-1.6\%}}& 8269.90 \textsubscript{\textcolor{ForestGreen}{{-0.1\%}}} & 14704.81\textsubscript{\textcolor{ForestGreen}{{-14.8\%}}} \\
    \bottomrule
    \end{tabular}
    }
     \end{table}
    \includegraphics[width=\linewidth]{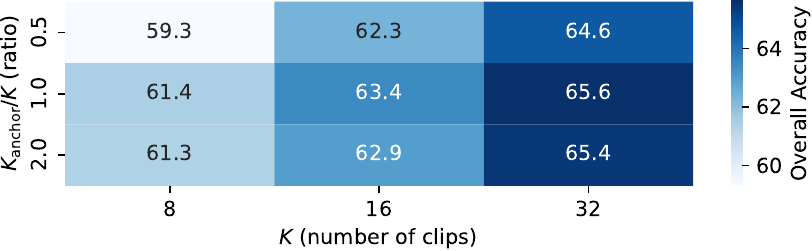}
    \vspace{-5mm}
    \caption{Performance across different $K$ and $K_\text{anchor}/K$ ratios on Video-MME.}
    \label{fig:k_ratio}
\end{minipage}
\vspace{-6mm}
\end{figure*}

\subsection{Experimental Settings}\label{sec:exp}

\noindent \textbf{Benchmarks.}  
We evaluate our method on three widely used long-form video benchmarks: Video-MME, LongVideoBench, and MLVU, which together cover diverse domains and evaluation protocols.
\textit{Video-MME}~\citep{fu2024video} is a large-scale multimodal evaluation suite designed to assess the capabilities of Video LLMs across video understanding, reasoning, and knowledge-grounded tasks. It includes both short and long-form videos, enabling a comprehensive evaluation of context management strategies.
\textit{LongVideoBench}~\citep{wu2024longvideobench} specifically targets long-form video understanding with a diverse set of tasks such as VQA, summarization, and temporal grounding. It emphasizes scenarios where videos span tens of minutes to hours, providing a challenging testbed for evaluating temporal reasoning and efficient context construction.
\textit{MLVU} (Multi-level Long Video Understanding)~\citep{zhou2024mlvu} is a benchmark focused on hierarchical understanding of long videos. It evaluates models on multiple levels of reasoning, including frame-level recognition, clip-level temporal understanding, and video-level holistic comprehension. This makes MLVU particularly suitable for testing whether methods like ours preserve both local and global context.

\noindent \textbf{Comparison Methods.}  
We compare our proposed method with several representative selection strategies that are applicable to long-form videos. \textit{Uniform sampling} selects frames at equal temporal intervals across the video. \textit{Top-$k$ sampling} selects the frames with the highest similarity scores to the given text query. The training-free method BOLT~\citep{BOLT} selects frames using Inverse Transform Sampling based on frame importance scores, AKS~\citep{aks} adaptively selects keyframes by recursively splitting video segments based on frame–text relevance, and Q-frame~\citep{zhang2025qframequeryawareframeselection} selects three levels of frames in different resolutions. 
In addition, we include watershed selection, which selects frames according to temporal boundaries detected by watershed segmentation. 

\noindent \textbf{Implementation Details.}  
We use Qwen2.5-VL-7B as the backbone VLM due to its support for any input resolution. All videos are loaded at 1 FPS for selection. For all baseline methods, we keep the original video resolution without resizing to ensure a fair comparison. Evaluations are conducted using the \texttt{lmm\_evals} library~\citep{zhang2024lmmsevalrealitycheckevaluation} on a computing cluster equipped with NVIDIA A100 GPUs. For similarity-based selection (Top-$k$), we use SigLIP2~\citep{tschannen2025siglip2multilingualvisionlanguage} as the vision-language model to compute similarity scores. More details are provided in the \textit{supplementary}~\ref{app:imp}.


\subsection{Results on Long-form Video Benchmarks}

Table~\ref{tab:main} summarizes the results on Video-MME, LongVideoBench, and MLVU under different frame budgets. We compare F2C with uniform sampling, Top-$k$, watershed, BOLT, AKS, and Q-Frame. 

Across all benchmarks, F2C consistently surpasses baselines. Compared with uniform sampling, it yields substantial improvements, with the largest gains on MLVU, followed by Video-MME, and smaller but steady gains on LongVideoBench. The benefit is most pronounced under small frame budgets (\eg, $K=8$), where uniform sampling often misses critical content, while F2C remains robust by preserving temporal continuity. As $K$ increases, the relative gap narrows, but F2C still retains a stable advantage. 

Compared with Q-Frame, which primarily adjusts resolution at the frame level, F2C extends selection to temporally coherent clips. This integration of adaptive resolution and temporal continuity provides richer context and consistently stronger results. Overall, these findings demonstrate that F2C offers a training-free and scalable solution for long-form video understanding.

\subsection{Results on Different Types of Questions}

To better understand F2C, we break down the results of Video-MME ($K=16$) into six question categories, as shown in Figure~\ref{fig:radar}, where we scale the range on each axis to 0.85$\times$ min to 1.01$\times$ max. Compared to uniform sampling, Top-$k$, watershed, BOLT, and Q-Frame, F2C achieves consistent improvements across all categories. The gains are especially large in \textit{Counting}, \textit{Recognition}, and \textit{Reasoning}, where temporal continuity and adaptive resolution provide richer motion and contextual cues. F2C also excels in \textit{Perception} and \textit{OCR}, indicating its ability to balance fine-grained detail with temporal coherence. 
Overall, these results demonstrate the benefits of F2C extend broadly across diverse question types, confirming its general effectiveness for long-form video understanding.

\section{Discussion}\label{sec:discussion}

\begin{figure*}[t]
\centering
    \includegraphics[width=\linewidth]{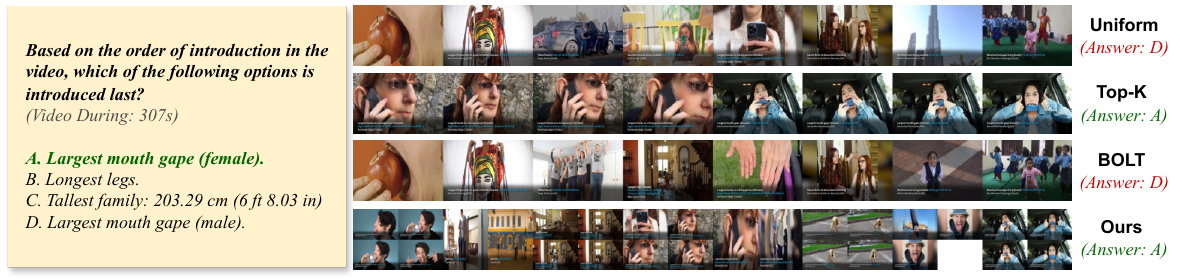}
    \vspace{-2mm}
    \caption{Visualization of the selected frames for each method.}
    \label{fig:vis}
    \vspace{-2mm}
\end{figure*}

\subsection{Ablation Study}

We conduct an ablation study on Video-MME with $K=16$ (Table~\ref{tab:ablation}), comparing uniform sampling, watershed selection, fixed-length key clips, and adaptive key clips. Watershed and key clip selection both outperform uniform sampling, showing the benefit of more informed context construction. Key clips are especially effective on long videos by preserving local temporal continuity, and adaptive resolution yields the best overall performance by balancing temporal coverage with spatial detail.

\subsection{Impact of Temporal Continuity}
\label{sec:temporal}

To examine whether temporal continuity provides greater benefit than simply sampling more frames, we conduct an experiment with a total of 64 frames. For baselines, 64 isolated frames are selected directly, while in F2C, the same budget is allocated to 16 clips with up to 4 frames each. To ensure fairness, all frames are downsampled with $s=2$, so the token count remains comparable across methods. Results are reported in Table~\ref{tab:temporal}.

F2C consistently outperforms the baselines, with the largest gain on long videos, where it reaches $54.1$ compared to $50.6$--$53.4$ for separate-frame approaches. This demonstrates that grouping frames into clips yields richer temporal cues than simply increasing frame count. 
Moreover, by introducing resolution as an additional factor in context management for long-form video understanding, such a resolution-aware design can further enhance existing frame selection methods.

\subsection{Discussion on Anchor Key Frame Selection}
\label{sec:initial}

\textbf{Selectors.} In F2C, we default to watershed for anchor keyframe selection, but other strategies can also be used. We apply F2C on top of uniform sampling, Top-$k$, BOLT, and watershed. As shown in Figure~\ref{fig:initial}, F2C consistently improves performance across all selectors, confirming the general benefit of extending anchors into clips. However, the quality of anchors matters: Top-$k$, which yields low-diversity frames, performs worst even after enhancement, while BOLT and watershed provide stronger results due to higher diversity. Uniform sampling benefits from diversity but lacks semantic relevance, limiting its effectiveness. These findings highlight that both diversity and relevance are crucial for anchor detection, and F2C effectively leverages them.

\noindent\textbf{Impact of $K_\text{anchor}$.} We further study the number of anchor frames by varying $K_\text{anchor}/K$, where $K$ is the total frame budget. Results in Figure~\ref{fig:k_ratio} show that smaller $K_\text{anchor}$ reduces diversity, while larger $K_\text{anchor}$ sacrifices temporal continuity. The best balance is achieved when $K_\text{anchor}=K$, which preserves both spatial detail and temporal coverage.

\subsection{Computational Analysis}
\label{sec:computation}

We analyze the computational efficiency of F2C by comparing visual token counts on Video-MME (w/o subtitles) under budgets of 8, 16, and 32 full-resolution frames (Table~\ref{tab:token_efficiency}). Despite adding temporal context, F2C achieves similar or fewer tokens than frame-based selectors because overlapping frames are encoded only once. The reduction becomes increasingly significant as the budget grows, showing that F2C not only enriches temporal continuity but also improves efficiency. 

It is also worth noting that the additional cost introduced by the vision-language encoder is minor relative to the overall VLM inference cost. First, image features can be pre-computed and cached, making the per-query overhead effectively zero during inference. Second, even when computed on the fly, the encoding is parallel across frames and can be executed in a single batch, whereas VLM decoding remains strictly sequential over tokens. As a result, the dominant computational bottleneck remains the LLM forward pass rather than the visual encoder. This balance of accuracy and computation further underscores the suitability of F2C for long-form video understanding.

\subsection{Visualization}

Figure~\ref{fig:vis} provides a qualitative comparison of different frame selection strategies on a long-form video. Uniform sampling and BOLT fail to capture the crucial frames corresponding to the correct answer, resulting in incorrect predictions. Top-K manages to select one relevant frame but lacks sufficient coverage of the question, leading to incomplete context. In contrast, our method successfully selects the key frames related to the correct option while also maintaining high diversity across the video, enabling more comprehensive temporal reasoning and accurate prediction. Besides, we provide a visualization on the distribution of the selections in the \textit{supplementary}~\ref{app:vis}.

\section{Conclusion}

We revisited the frame selection for long-form video understanding and proposed Frames-to-Clips, a training-free framework that replaces isolated keyframes with temporally coherent clips and introduces an adaptive trade-off between resolution and clip length under a fixed token budget. Extensive experiments on Video-MME, LongVideoBench, and MLVU show that F2C consistently outperforms state-of-the-art training-free baselines. Our analyses further demonstrate that temporal continuity, diversity in anchor selection, and balanced resolution scaling are all crucial for effective context management. By improving selection without training, F2C offers a simple and scalable solution for long-form video understanding and points toward future directions for VLMs.

{
    \small
    \bibliographystyle{ieeenat_fullname}
    \bibliography{main}

@String(CVPR= {IEEE Conf. Comput. Vis. Pattern Recog.})

@String(ECCV= {Eur. Conf. Comput. Vis.})

@String(CVPR  = {CVPR})

@String(ECCV  = {ECCV})

@article{fu2024video,
  title={Video-MME: The First-Ever Comprehensive Evaluation Benchmark of Multi-modal LLMs in Video Analysis},
  author={Fu, Chaoyou and Dai, Yuhan and Luo, Yondong and Li, Lei and Ren, Shuhuai and Zhang, Renrui and Wang, Zihan and Zhou, Chenyu and Shen, Yunhang and Zhang, Mengdan and others},
  journal={arXiv preprint arXiv:2405.21075},
  year={2024}
}

@inproceedings{radford2021learning,
  title={Learning transferable visual models from natural language supervision},
  author={Radford, Alec and Kim, Jong Wook and Hallacy, Chris and Ramesh, Aditya and Goh, Gabriel and Agarwal, Sandhini and Sastry, Girish and Askell, Amanda and Mishkin, Pamela and Clark, Jack and others},
  booktitle={International conference on machine learning},
  pages={8748--8763},
  year={2021},
  organization={PMLR}
}

@inproceedings{chenShareGPT4VideoImprovingVideo2024,
  title = {ShareGPT4Video: Improving Video Understanding and Generation with Better Captions},
  author={Chen, Lin and Wei, Xilin and Li, Jinsong and Dong, Xiaoyi and Zhang, Pan and Zang, Yuhang and Chen, Zehui and Duan, Haodong and Tang, Zhenyu and Yuan, Li and others},
  booktitle={NeurIPS},
  volume={37},
  pages={19472--19495},
  year={2024}
}

@inproceedings{wuDIBSEnhancingDense2024,
  title = {DIBS: Enhancing Dense Video Captioning with Unlabeled Videos via Pseudo Boundary Enrichment and Online Refinement},
  shorttitle = {DIBS},
  booktitle = {2024 IEEE/CVF Conference on Computer Vision and Pattern Recognition (CVPR)},
  author = {Wu, Hao and Liu, Huabin and Qiao, Yu and Sun, Xiao},
  year = {2024},
  month = jun,
  pages = {18699--18708},
  issn = {2575-7075}
}

@article{maaz2023video,
  title={Video-chatgpt: Towards detailed video understanding via large vision and language models},
  author={Maaz, Muhammad and Rasheed, Hanoona and Khan, Salman and Khan, Fahad Shahbaz},
  journal={arXiv preprint arXiv:2306.05424},
  year={2023}
}

@article{lin2023mm,
  title={Mm-vid: Advancing video understanding with gpt-4v (ision)},
  author={Lin, Kevin and Ahmed, Faisal and Li, Linjie and Lin, Chung-Ching and Azarnasab, Ehsan and Yang, Zhengyuan and Wang, Jianfeng and Liang, Lin and Liu, Zicheng and Lu, Yumao and others},
  journal={arXiv preprint arXiv:2310.19773},
  year={2023}
}

@article{zhang2023simple,
  title={A simple llm framework for long-range video question-answering},
  author={Zhang, Ce and Lu, Taixi and Islam, Md Mohaiminul and Wang, Ziyang and Yu, Shoubin and Bansal, Mohit and Bertasius, Gedas},
  journal={arXiv preprint arXiv:2312.17235},
  year={2023}
}

@article{lin2023video,
  title={Video-llava: Learning united visual representation by alignment before projection},
  author={Lin, Bin and Ye, Yang and Zhu, Bin and Cui, Jiaxi and Ning, Munan and Jin, Peng and Yuan, Li},
  journal={arXiv preprint arXiv:2311.10122},
  year={2023}
}

@inproceedings{jin2024chat,
  title={Chat-univi: Unified visual representation empowers large language models with image and video understanding},
  author={Jin, Peng and Takanobu, Ryuichi and Zhang, Wancai and Cao, Xiaochun and Yuan, Li},
  booktitle={CVPR},
  pages={13700--13710},
  year={2024}
}

@inproceedings{ren2024timechat,
  title={Timechat: A time-sensitive multimodal large language model for long video understanding},
  author={Ren, Shuhuai and Yao, Linli and Li, Shicheng and Sun, Xu and Hou, Lu},
  booktitle={Proceedings of the IEEE/CVF Conference on Computer Vision and Pattern Recognition},
  pages={14313--14323},
  year={2024}
}

@misc{yangVid2SeqLargeScalePretraining2023,
  title = {Vid2Seq: Large-Scale Pretraining of a Visual Language Model for Dense Video Captioning},
  shorttitle = {Vid2Seq},
  author = {Yang, Antoine and Nagrani, Arsha and Seo, Paul Hongsuck and Miech, Antoine and {Pont-Tuset}, Jordi and Laptev, Ivan and Sivic, Josef and Schmid, Cordelia},
  year = {2023},
  month = mar,
  number = {arXiv:2302.14115},
  publisher = {arXiv},
  archiveprefix = {arXiv}
}

@article{kimImageGridCan2024,
  title={An image grid can be worth a video: Zero-shot video question answering using a vlm},
  author={Kim, Wonkyun and Choi, Changin and Lee, Wonseok and Rhee, Wonjong},
  journal={IEEE Access},
  year={2024},
  publisher={IEEE}
}

@inproceedings{minMoReVQAExploringModular2024,
  title = {MoReVQA: Exploring Modular Reasoning Models for Video Question Answering},
  shorttitle = {MoReVQA},
  booktitle = {2024 IEEE/CVF Conference on Computer Vision and Pattern Recognition (CVPR)},
  author = {Min, Juhong and Buch, Shyamal and Nagrani, Arsha and Cho, Minsu and Schmid, Cordelia},
  year = {2024},
  month = jun,
  pages = {13235--13245},
  issn = {2575-7075}
}

@misc{liLLaVAOneVisionEasyVisual2024a,
  title = {LLaVA-OneVision: Easy Visual Task Transfer},
  shorttitle = {LLaVA-OneVision},
  author = {Li, Bo and Zhang, Yuanhan and Guo, Dong and Zhang, Renrui and Li, Feng and Zhang, Hao and Zhang, Kaichen and Zhang, Peiyuan and Li, Yanwei and Liu, Ziwei and Li, Chunyuan},
  year = {2024},
  month = oct,
  number = {arXiv:2408.03326},
  publisher = {arXiv},
  archiveprefix = {arXiv}
}

@article{li2023videochat,
  title={Videochat: Chat-centric video understanding},
  author={Li, KunChang and He, Yinan and Wang, Yi and Li, Yizhuo and Wang, Wenhai and Luo, Ping and Wang, Yali and Wang, Limin and Qiao, Yu},
  journal={arXiv preprint arXiv:2305.06355},
  year={2023}
}

@inproceedings{ataallahGoldfishVisionLanguageUnderstanding2024a,
  title = {Goldfish: Vision-Language Understanding of Arbitrarily Long Videos},
  author={Ataallah, Kirolos and Shen, Xiaoqian and Abdelrahman, Eslam and Sleiman, Essam and Zhuge, Mingchen and Ding, Jian and Zhu, Deyao and Schmidhuber, J{\"u}rgen and Elhoseiny, Mohamed},
  booktitle={ECCV},
  pages={251--267},
  year={2024},
  organization={Springer}
}

@misc{yuFrameVoyagerLearningQuery2024,
  title = {Frame-Voyager: Learning to Query Frames for Video Large Language Models},
  shorttitle = {Frame-Voyager},
  author = {Yu, Sicheng and Jin, Chengkai and Wang, Huanyu and Chen, Zhenghao and Jin, Sheng and Zuo, Zhongrong and Xu, Xiaolei and Sun, Zhenbang and Zhang, Bingni and Wu, Jiawei and Zhang, Hao and Sun, Qianru},
  year = {2024},
  month = oct,
  number = {arXiv:2410.03226},
  publisher = {arXiv},
  archiveprefix = {arXiv}
}

@misc{shenLongVUSpatiotemporalAdaptive2024,
  title = {LongVU: Spatiotemporal Adaptive Compression for Long Video-Language Understanding},
  shorttitle = {LongVU},
  author = {Shen, Xiaoqian and Xiong, Yunyang and Zhao, Changsheng and Wu, Lemeng and Chen, Jun and Zhu, Chenchen and Liu, Zechun and Xiao, Fanyi and Varadarajan, Balakrishnan and Bordes, Florian and Liu, Zhuang and Xu, Hu and Kim, Hyunwoo J. and Soran, Bilge and Krishnamoorthi, Raghuraman and Elhoseiny, Mohamed and Chandra, Vikas},
  year = {2024},
  month = oct,
  number = {arXiv:2410.17434},
  publisher = {arXiv},
  archiveprefix = {arXiv}
}

@misc{baiQwenVLVersatileVisionLanguage2023,
  title = {Qwen-{{VL}}: {{A Versatile Vision-Language Model}} for {{Understanding}}, {{Localization}}, {{Text Reading}}, and {{Beyond}}},
  shorttitle = {Qwen-{{VL}}},
  author = {Bai, Jinze and Bai, Shuai and Yang, Shusheng and Wang, Shijie and Tan, Sinan and Wang, Peng and Lin, Junyang and Zhou, Chang and Zhou, Jingren},
  year = {2023},
  month = oct,
  number = {arXiv:2308.12966},
  eprint = {2308.12966},
  publisher = {arXiv},
  doi = {10.48550/arXiv.2308.12966}
}

@misc{liu2024llavanext,
    title={LLaVA-NeXT: Improved reasoning, OCR, and world knowledge},
    author={Liu, Haotian and Li, Chunyuan and Li, Yuheng and Li, Bo and Zhang, Yuanhan and Shen, Sheng and Lee, Yong Jae},
    month={January},
    year={2024}
}

@misc{chengVideoLLaMA2Advancing2024,
  title = {{{VideoLLaMA}} 2: {{Advancing Spatial-Temporal Modeling}} and {{Audio Understanding}} in {{Video-LLMs}}},
  shorttitle = {{{VideoLLaMA}} 2},
  author = {Cheng, Zesen and Leng, Sicong and Zhang, Hang and Xin, Yifei and Li, Xin and Chen, Guanzheng and Zhu, Yongxin and Zhang, Wenqi and Luo, Ziyang and Zhao, Deli and Bing, Lidong},
  year = {2024},
  month = oct,
  number = {arXiv:2406.07476},
  eprint = {2406.07476},
  publisher = {arXiv},
  doi = {10.48550/arXiv.2406.07476}
}

@misc{xueLongVILAScalingLongContext2024,
  title = {{{LongVILA}}: {{Scaling Long-Context Visual Language Models}} for {{Long Videos}}},
  shorttitle = {{{LongVILA}}},
  author = {Xue, Fuzhao and Chen, Yukang and Li, Dacheng and Hu, Qinghao and Zhu, Ligeng and Li, Xiuyu and Fang, Yunhao and Tang, Haotian and Yang, Shang and Liu, Zhijian and He, Ethan and Yin, Hongxu and Molchanov, Pavlo and Kautz, Jan and Fan, Linxi and Zhu, Yuke and Lu, Yao and Han, Song},
  year = {2024},
  month = nov,
  number = {arXiv:2408.10188},
  eprint = {2408.10188},
  publisher = {arXiv},
  doi = {10.48550/arXiv.2408.10188}
}

@misc{shuVideoXLExtraLongVision2024,
  title = {Video-{{XL}}: {{Extra-Long Vision Language Model}} for {{Hour-Scale Video Understanding}}},
  shorttitle = {Video-{{XL}}},
  author = {Shu, Yan and Zhang, Peitian and Liu, Zheng and Qin, Minghao and Zhou, Junjie and Huang, Tiejun and Zhao, Bo},
  year = {2024},
  month = oct,
  number = {arXiv:2409.14485},
  eprint = {2409.14485},
  publisher = {arXiv},
  doi = {10.48550/arXiv.2409.14485},
  urldate = {2024-11-14}
}

@misc{zhang2024lmmsevalrealitycheckevaluation,
      title={LMMs-Eval: Reality Check on the Evaluation of Large Multimodal Models}, 
      author={Kaichen Zhang and Bo Li and Peiyuan Zhang and Fanyi Pu and Joshua Adrian Cahyono and Kairui Hu and Shuai Liu and Yuanhan Zhang and Jingkang Yang and Chunyuan Li and Ziwei Liu},
      year={2024},
      eprint={2407.12772},
      archivePrefix={arXiv},
      primaryClass={cs.CL},
}

@inproceedings{zhai2023sigmoid,
  title={Sigmoid loss for language image pre-training},
  author={Zhai, Xiaohua and Mustafa, Basil and Kolesnikov, Alexander and Beyer, Lucas},
  booktitle={Proceedings of the IEEE/CVF International Conference on Computer Vision},
  pages={11975--11986},
  year={2023}
}

@article{zhou2024mlvu,
  title={Mlvu: A comprehensive benchmark for multi-task long video understanding},
  author={Zhou, Junjie and Shu, Yan and Zhao, Bo and Wu, Boya and Xiao, Shitao and Yang, Xi and Xiong, Yongping and Zhang, Bo and Huang, Tiejun and Liu, Zheng},
  journal={arXiv preprint arXiv:2406.04264},
  year={2024}
}

@InProceedings{haralick1987image,
  title={Image analysis using mathematical morphology},
  author={Haralick, Robert M and Sternberg, Stanley R and Zhuang, Xinhua},
  booktitle={TPAMI},
  number={4},
  volume={1},
  pages={532--550},
  year={1987},
  publisher={IEEE}
}

@article{wu2024longvideobench,
  title={Longvideobench: A benchmark for long-context interleaved video-language understanding},
  author={Wu, Haoning and Li, Dongxu and Chen, Bei and Li, Junnan},
  journal={NeurIPS},
  volume={37},
  pages={28828--28857},
  year={2024}
}

@InProceedings{Liu_2024_CVPR,
    author    = {Liu, Shuming and Zhang, Chen-Lin and Zhao, Chen and Ghanem, Bernard},
    title     = {End-to-End Temporal Action Detection with 1B Parameters Across 1000 Frames},
    booktitle = {CVPR},
    month     = {June},
    year      = {2024},
    pages     = {18591-18601}
}

@misc{bai2025qwen25vltechnicalreport,
      title={Qwen2.5-VL Technical Report}, 
      author={Shuai Bai and Keqin Chen and Xuejing Liu and Jialin Wang and Wenbin Ge and Sibo Song and Kai Dang and Peng Wang and Shijie Wang and Jun Tang and Humen Zhong and Yuanzhi Zhu and Mingkun Yang and Zhaohai Li and Jianqiang Wan and Pengfei Wang and Wei Ding and Zheren Fu and Yiheng Xu and Jiabo Ye and Xi Zhang and Tianbao Xie and Zesen Cheng and Hang Zhang and Zhibo Yang and Haiyang Xu and Junyang Lin},
      year={2025},
      eprint={2502.13923},
      archivePrefix={arXiv},
      primaryClass={cs.CV},
}

@INPROCEEDINGS {BOLT,
author = { Liu, Shuming and Zhao, Chen and Xu, Tianqi and Ghanem, Bernard },
booktitle = { 2025 IEEE/CVF Conference on Computer Vision and Pattern Recognition (CVPR) },
title = {{ BOLT: Boost Large Vision-Language Model Without Training for Long-Form Video Understanding }},
year = {2025},
volume = {},
ISSN = {},
pages = {3318-3327},
doi = {10.1109/CVPR52734.2025.00315},
publisher = {IEEE Computer Society},
address = {Los Alamitos, CA, USA},
month =Jun}

@misc{tstar,
title={Re-thinking Temporal Search for Long-Form Video Understanding}, 
author={Jinhui Ye and Zihan Wang and Haosen Sun and Keshigeyan Chandrasegaran and Zane Durante and Cristobal Eyzaguirre and Yonatan Bisk and Juan Carlos Niebles and Ehsan Adeli and Li Fei-Fei and Jiajun Wu and Manling Li},
year={2025},
eprint={2504.02259},
archivePrefix={arXiv},
primaryClass={cs.CV},
}

@misc{tschannen2025siglip2multilingualvisionlanguage,
      title={SigLIP 2: Multilingual Vision-Language Encoders with Improved Semantic Understanding, Localization, and Dense Features}, 
      author={Michael Tschannen and Alexey Gritsenko and Xiao Wang and Muhammad Ferjad Naeem and Ibrahim Alabdulmohsin and Nikhil Parthasarathy and Talfan Evans and Lucas Beyer and Ye Xia and Basil Mustafa and Olivier Hénaff and Jeremiah Harmsen and Andreas Steiner and Xiaohua Zhai},
      year={2025},
      eprint={2502.14786},
      archivePrefix={arXiv},
      primaryClass={cs.CV},
}

@Article{Hu2025,
 author = {Kai Hu and Feng Gao and Xiaohan Nie and Peng Zhou and Son Tran and Tal Neiman and Lingyun Wang and Mubarak Shah and Raffay Hamid and Bing Yin and Trishul Chilimbi},
 title = {M-LLM based video frame selection for efficient video understanding},
 year = {2025},
    booktitle = { 2025 IEEE/CVF Conference on Computer Vision and Pattern Recognition (CVPR) }

}

@article{yao2025generative,
    title={Generative Frame Sampler for Long Video Understanding},
    author={Yao, Linli and Wu, Haoning and Ouyang, Kun and Zhang, Yuanxing and Xiong, Caiming and Chen, Bei and Sun, Xu and Li, Junnan},
    journal={arXiv preprint arXiv:2503.09146},
    year={2025}
}

@inproceedings{yu2023self,
  title   = {Self-Chained Image-Language Model for Video Localization and Question Answering},
  author  = {Yu, Shoubin and Cho, Jaemin and Yadav, Prateek and Bansal, Mohit},
  booktitle = {NeurIPS},
  year    = {2023}
}

@misc{fei2024videoccamenhancingvideolanguageunderstanding,
      title={Video-CCAM: Enhancing Video-Language Understanding with Causal Cross-Attention Masks for Short and Long Videos}, 
      author={Jiajun Fei and Dian Li and Zhidong Deng and Zekun Wang and Gang Liu and Hui Wang},
      year={2024},
      eprint={2408.14023},
      archivePrefix={arXiv},
      primaryClass={cs.CV},
}

@article{zhang2024longva,
  title={Long Context Transfer from Language to Vision},
  author={Peiyuan Zhang and Kaichen Zhang and Bo Li and Guangtao Zeng and Jingkang Yang and Yuanhan Zhang and Ziyue Wang and Haoran Tan and Chunyuan Li and Ziwei Liu},
  journal={arXiv preprint arXiv:2406.16852},
  year={2024},
}

@misc{sun2025mdp3trainingfreeapproachlistwise,
      title={MDP3: A Training-free Approach for List-wise Frame Selection in Video-LLMs}, 
      author={Hui Sun and Shiyin Lu and Huanyu Wang and Qing-Guo Chen and Zhao Xu and Weihua Luo and Kaifu Zhang and Ming Li},
      year={2025},
      eprint={2501.02885},
      archivePrefix={arXiv},
      primaryClass={cs.CV},
}

@misc{buch2022revisitingvideovideolanguageunderstanding,
      title={Revisiting the "Video" in Video-Language Understanding}, 
      author={Shyamal Buch and Cristóbal Eyzaguirre and Adrien Gaidon and Jiajun Wu and Li Fei-Fei and Juan Carlos Niebles},
      year={2022},
      eprint={2206.01720},
      archivePrefix={arXiv},
      primaryClass={cs.CV},
}

@inproceedings{buch2025flexible,
  title={Flexible Frame Selection for Efficient Video Reasoning},
  author={Shyamal Buch and Arsha Nagrani and Anurag Arnab and Cordelia Schmid},
  booktitle={CVPR},
  year={2025}
}

@misc{xu2025viarladaptivetemporalgrounding,
      title={ViaRL: Adaptive Temporal Grounding via Visual Iterated Amplification Reinforcement Learning}, 
      author={Ziqiang Xu and Qi Dai and Tian Xie and Yifan Yang and Kai Qiu and DongDong Chen and Zuxuan Wu and Chong Luo},
      year={2025},
      eprint={2505.15447},
      archivePrefix={arXiv},
      primaryClass={cs.CV},
}

@misc{ghazanfari2025chainofframesadvancingvideounderstanding,
      title={Chain-of-Frames: Advancing Video Understanding in Multimodal LLMs via Frame-Aware Reasoning}, 
      author={Sara Ghazanfari and Francesco Croce and Nicolas Flammarion and Prashanth Krishnamurthy and Farshad Khorrami and Siddharth Garg},
      year={2025},
      eprint={2506.00318},
      archivePrefix={arXiv},
      primaryClass={cs.CV},
}

@misc{arnab2025temporal,
    title={Temporal Chain of Thought: Long-Video Understanding by Thinking in Frames},
    author={Anurag Arnab and Ahmet Iscen and Mathilde Caron and Alireza Fathi and Cordelia Schmid},
    year={2025},
    eprint={2507.02001},
    archivePrefix={arXiv},
    primaryClass={cs.LG}
}

@misc{geminiteam2024gemini15unlockingmultimodal,
      title={Gemini 1.5: Unlocking multimodal understanding across millions of tokens of context}, 
      author={Gemini Team and Petko Georgiev and Ving Ian Lei et al.},
      year={2024},
      eprint={2403.05530},
      archivePrefix={arXiv},
      primaryClass={cs.CL},
}

@misc{zhang2025qframequeryawareframeselection,
      title={Q-Frame: Query-aware Frame Selection and Multi-Resolution Adaptation for Video-LLMs}, 
      author={Shaojie Zhang and Jiahui Yang and Jianqin Yin and Zhenbo Luo and Jian Luan},
      year={2025},
      eprint={2506.22139},
      archivePrefix={arXiv},
      primaryClass={cs.CV},
}

@misc{lee2025refocusreinforcementguidedframeoptimization,
      title={ReFoCUS: Reinforcement-guided Frame Optimization for Contextual Understanding}, 
      author={Hosu Lee and Junho Kim and Hyunjun Kim and Yong Man Ro},
      year={2025},
      eprint={2506.01274},
      archivePrefix={arXiv},
      primaryClass={cs.CV},
}

@misc{shangguan2025tomatoassessingvisualtemporal,
      title={TOMATO: Assessing Visual Temporal Reasoning Capabilities in Multimodal Foundation Models}, 
      author={Ziyao Shangguan and Chuhan Li and Yuxuan Ding and Yanan Zheng and Yilun Zhao and Tesca Fitzgerald and Arman Cohan},
      year={2025},
      eprint={2410.23266},
      archivePrefix={arXiv},
      primaryClass={cs.CV},
}

@article{zhang2024longclip,
        title={Long-CLIP: Unlocking the Long-Text Capability of CLIP},
        author={Beichen Zhang and Pan Zhang and Xiaoyi Dong and Yuhang Zang and Jiaqi Wang},
        journal={arXiv preprint arXiv:2403.15378},
        year={2024}
}

@INPROCEEDINGS {aks,
author = { Tang, Xi and Qiu, Jihao and Xie, Lingxi and Tian, Yunjie and Jiao, Jianbin and Ye, Qixiang },
booktitle = { 2025 IEEE/CVF Conference on Computer Vision and Pattern Recognition (CVPR) },
title = {{ Adaptive Keyframe Sampling for Long Video Understanding }},
year = {2025},
volume = {},
ISSN = {},
pages = {29118-29128},
publisher = {IEEE Computer Society},
address = {Los Alamitos, CA, USA},
month =Jun}

@misc{keye,
      title={Kwai Keye-VL 1.5 Technical Report}, 
      author={Kwai Keye Team},
      year={2025},
      eprint={2509.01563},
      archivePrefix={arXiv},
      primaryClass={cs.CV},
      url={https://arxiv.org/abs/2509.01563}, 
}

@article{vilamp,
  title={Scaling Video-Language Models to 10K Frames via Hierarchical Differential Distillation},
  author={Cheng, Chuanqi and Guan, Jian and Wu, Wei and Yan, Rui},
  journal={arXiv preprint arXiv:2504.02438},
  year={2025}
}

@misc{xu2025slowfastllava15familytokenefficientvideo,
      title={SlowFast-LLaVA-1.5: A Family of Token-Efficient Video Large Language Models for Long-Form Video Understanding}, 
      author={Mingze Xu and Mingfei Gao and Shiyu Li and Jiasen Lu and Zhe Gan and Zhengfeng Lai and Meng Cao and Kai Kang and Yinfei Yang and Afshin Dehghan},
      year={2025},
      eprint={2503.18943},
      archivePrefix={arXiv},
      primaryClass={cs.CV},
      url={https://arxiv.org/abs/2503.18943}, 
}

@article{xu2024slowfast,
  title={Slowfast-llava: A strong training-free baseline for video large language models},
  author={Xu, Mingze and Gao, Mingfei and Gan, Zhe and Chen, Hong-You and Lai, Zhengfeng and Gang, Haiming and Kang, Kai and Dehghan, Afshin},
  journal={arXiv preprint arXiv:2407.15841},
  year={2024}
}

@article{song2025moviechat,
  title={Moviechat+: Question-aware sparse memory for long video question answering},
  author={Song, Enxin and Chai, Wenhao and Ye, Tian and Hwang, Jenq-Neng and Li, Xi and Wang, Gaoang},
  journal={IEEE Transactions on Pattern Analysis and Machine Intelligence},
  year={2025},
  publisher={IEEE}
}

@article{dos2025video,
  title={Infinity-Video: A Training-Free Approach to Long Video Understanding via Continuous-Time Memory Consolidation},
  author={dos Santos, Saul Jos{\'e} Rodrigues and Farinhas, Ant{\'o}nio and McNamee, Daniel C and Martins, Andr{\'e} FT},
  journal={CoRR},
  year={2025}
}

@misc{hu2025coschainofshotpromptinglong,
      title={CoS: Chain-of-Shot Prompting for Long Video Understanding}, 
      author={Jian Hu and Zixu Cheng and Chenyang Si and Wei Li and Shaogang Gong},
      year={2025},
      eprint={2502.06428},
      archivePrefix={arXiv},
      primaryClass={cs.CV},
      url={https://arxiv.org/abs/2502.06428}, 
}
}

\clearpage
\setcounter{page}{1}
\appendix
\maketitlesupplementary

\section{Overview}
This appendix provides additional details and supporting analyses to complement the main paper. Specifically, we include:
\begin{itemize}[leftmargin=*, topsep=0pt, itemsep=1pt, partopsep=0pt, parsep=0pt]
    \item \textbf{Usage of Large Language Models} (Section~\ref{app:llm}): clarifies how LLMs were used to assist in writing and formatting.
    \item \textbf{Limitations} (Section~\ref{app:limit}): discusses constraints of F2C and dependency on backbone VLMs.
    \item \textbf{Implementation Details} (Section~\ref{app:imp}): provides prompts, pseudo-code for watershed selection, and re-implementation details of BOLT and Q-Frame.
    \item \textbf{Visualization} (Section~\ref{app:vis}): illustrates qualitative comparisons of frame selection strategies and cases analysis on F2C.
    \item \textbf{Additional Experiments} (Section~\ref{app:clip}): analyzes the impact of different CLIP backbones on performance.
\end{itemize}


\section{Usage of Large Language Model}\label{app:llm}

We used a Large Language Model (LLM) to assist in polishing the writing style and generating \LaTeX{} code for tables. All technical ideas, experimental designs, and analyses were developed by the authors.

\section{Limitation}\label{app:limit}

Although F2C effectively improves context management by providing temporally coherent and resolution-adaptive inputs, its performance upper bound is still constrained by the capability of the downstream VLM. 
Even when the correct frames or clips are selected, the final answer depends on the reasoning and comprehension ability of the VLM itself. As a result, limitations in temporal reasoning, spatial understanding, or multimodal alignment within the backbone model remain bottlenecks. Our method is thus complementary to future advances in Video LLM architectures, and its benefits may further amplify as more powerful backbone models become available. A further discussion and visualization are provided in Section~\ref{app:vis_f2c}.

\section{Implementation Details}\label{app:imp}

\subsection{Prompt for VQA}

We use the following prompt to query VLM for the VQA task with temperature of 0.0 for reproducibility.

\begin{promptbox}
    {Prompt for Video Question Answering\label{prompt}}{
    $<$Selected Frames$>$
    \\\\
    Select the best answer to the following multiple-choice question based on the video and the subtitles. Respond with only the letter (A, B, C, or D) of the correct option.
    \\\\
    $<$Questions$>$\\\\
    $<$Options$>$
\\\\
    Answer with the option's letter from the given choices directly.}
\end{promptbox}

We use the default random seeds from \texttt{lmms\_eval} for \texttt{random}, \texttt{numpy}, and \texttt{torch} as $0, 1234,$ and $1234$, respectively. 

\subsection{Pseudo-code of Watershed Selection}\label{app:water}

We provide the pseudo-code of our watershed-based anchor keyframe selection strategy in Algorithm~\ref{alg:watershed}. This procedure identifies local maxima of frame–text similarity within basins, then clusters them if necessary to ensure both diversity and relevance of the selected anchors.

\subsection{Hyperparameter of F2C}

We choose $\lambda_r=0.5$ and $\lambda_l=0.05$ for the adaptive clip length selection and set $s_\text{max}=2$ and $K_\text{anchor}=K$ for the experiments.

\begin{algorithm}[t]
\caption{Watershed Selection}
\label{alg:watershed}
\DontPrintSemicolon  
\SetAlgoNoLine      

\KwIn{Frame features $\{f_i\}_{i=1}^N$, text feature $q$, number of anchors $K_\text{anchor}$}
\KwOut{Indices of selected anchor frames}

- Normalize frame and text features, compute similarities 

\hspace{0.5cm} $s_i = \cos(f_i, q)$. \\

- Find valleys (local minima) in the similarity curve to define basin boundaries. \\

- For each basin, identify the peak frame (highest similarity) as a candidate. \\

- \textbf{If} number of candidates $>$ $K_\text{anchor}$: \\
\hspace{0.5cm} Cluster candidates into $K_\text{anchor}$ groups (via k-means). \\
\hspace{0.5cm} In each cluster, select the frame with the highest similarity. \\

- \textbf{Else:} \\
\hspace{0.5cm} Use all candidates (up to $K_\text{anchor}$). \\

\textbf{Return} the sorted list of selected anchor frames.  
\end{algorithm}

\subsection{Re-implementation of BOLT, Q-Frame, and AKS}

For a fair comparison, we closely followed the official settings reported in the original papers. 
For BOLT~\citep{BOLT}, we adopted the same hyperparameters as ours and set $\alpha=2.5$ as recommended in their work. 
For Q-Frame~\citep{zhang2025qframequeryawareframeselection}, the original implementation evaluates three configurations of $(\text{num}_{\text{high}}, \text{num}_{\text{medium}}, \text{num}_{\text{low}})=(4,8,32)$, which corresponds to our 8-frame setting. 
To extend Q-Frame under larger budgets, we applied proportional scaling and used $(8,16,64)$ for the 16-frame setting and $(16,32,128)$ for the 32-frame setting.
For AKS~\citep{aks}, we integrate the official implementation (\texttt{t1}=$0.1$, \texttt{t2}=$-100$, and \texttt{all\_depth}=5) into our framework with the same SigLIP2 feature extractor as other methods for a fair comparison on frame selection.
These adjustments ensure that both baselines are re-implemented consistently and evaluated within our experimental framework, facilitating fair comparison and reproducibility.

\section{Visualization} \label{app:vis}

\subsection{Distribution of Selection}
\begin{figure*}[t]
\centering
    \includegraphics[width=0.8\linewidth]{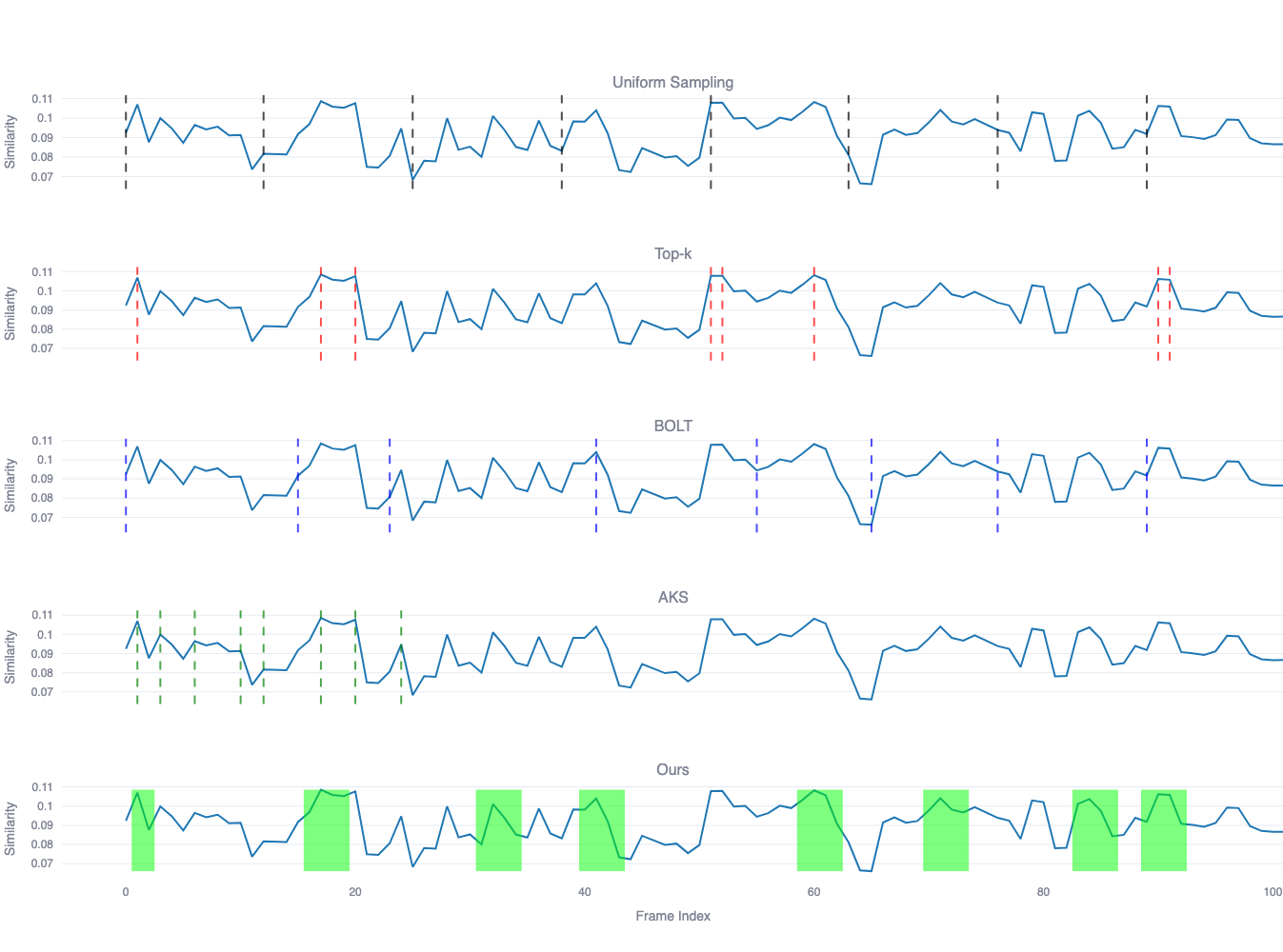}
     \includegraphics[width=0.8\linewidth]{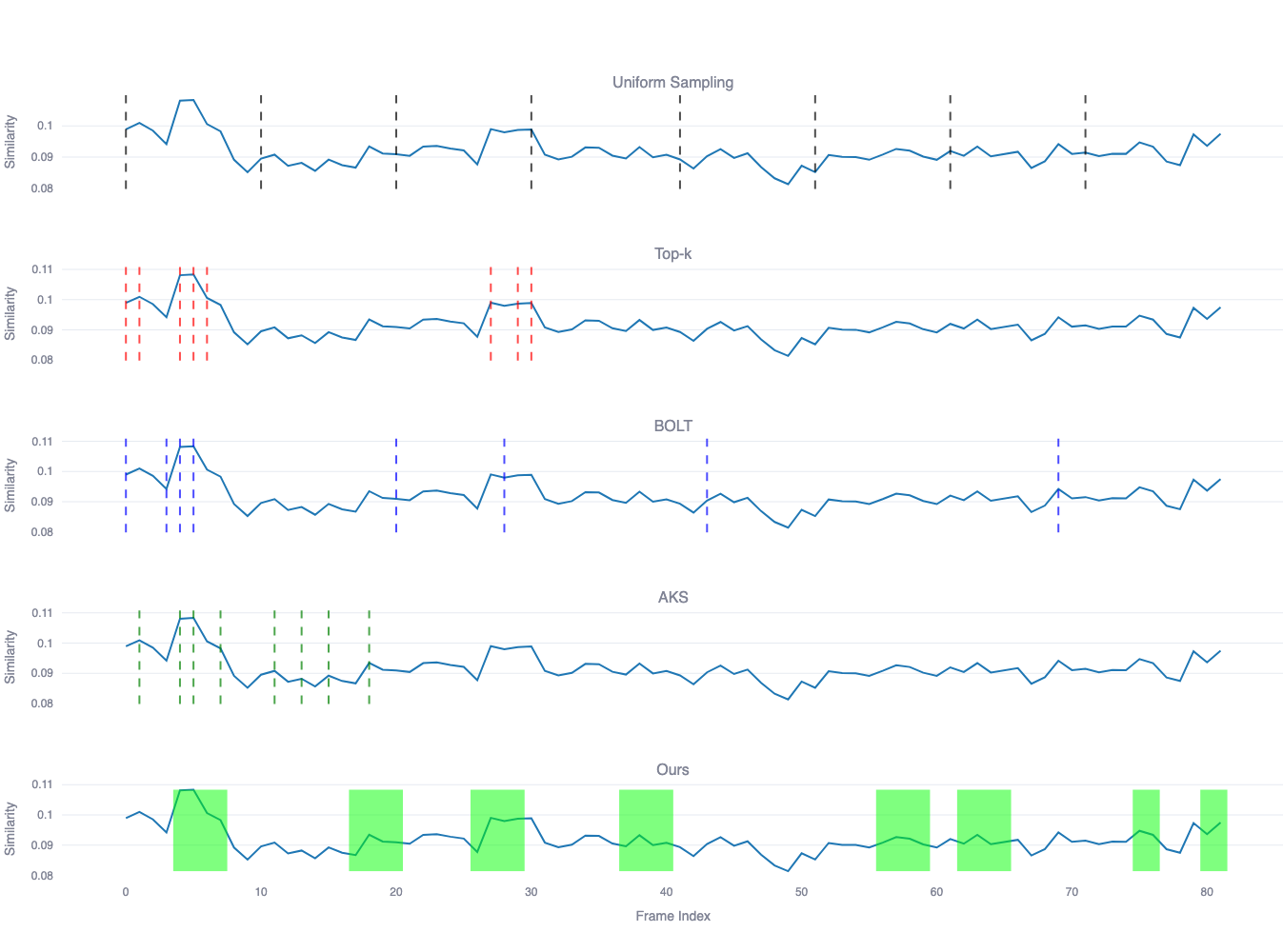}
    \caption{\textbf{Visualization of frame selection distributions for different methods.}
The similarity curve denotes the frame-level relevance to the query. Uniform sampling distributes frames evenly but ignores semantic peaks; BOLT concentrates on high-similarity regions yet exhibits redundancy; Top-$k$ and AKS select excessively clustered frames with limited coverage. In contrast, F2C selects temporally coherent clips that cover diverse, high-relevance regions while maintaining temporal continuity, leading to more informative and efficient context construction.}
    \label{fig:dist}
\end{figure*}

To complement the qualitative examples in the main paper, we provide an additional visualization of how different methods distribute their selected frames across the video (Figure~\ref{fig:dist}). The similarity curve represents the frame-level relevance to the query. 

Uniform sampling spreads frames evenly but ignores semantic signals, often missing important peaks. BOLT tends to favor high-similarity regions but suffers from redundancy. Top-$k$ and AKS focus too narrowly, selecting clustered frames around peaks while overlooking other segments. In contrast, F2C selects temporally coherent clips that cover diverse regions with high relevance, striking a better balance between coverage and precision. 

This visualization further illustrates that F2C not only captures key peaks but also preserves temporal continuity, leading to more informative and efficient context construction. 

\subsection{Selected Frames of F2C} \label{app:vis_f2c}

\begin{figure*}[t]
    \centering
    \includegraphics[width=\linewidth]{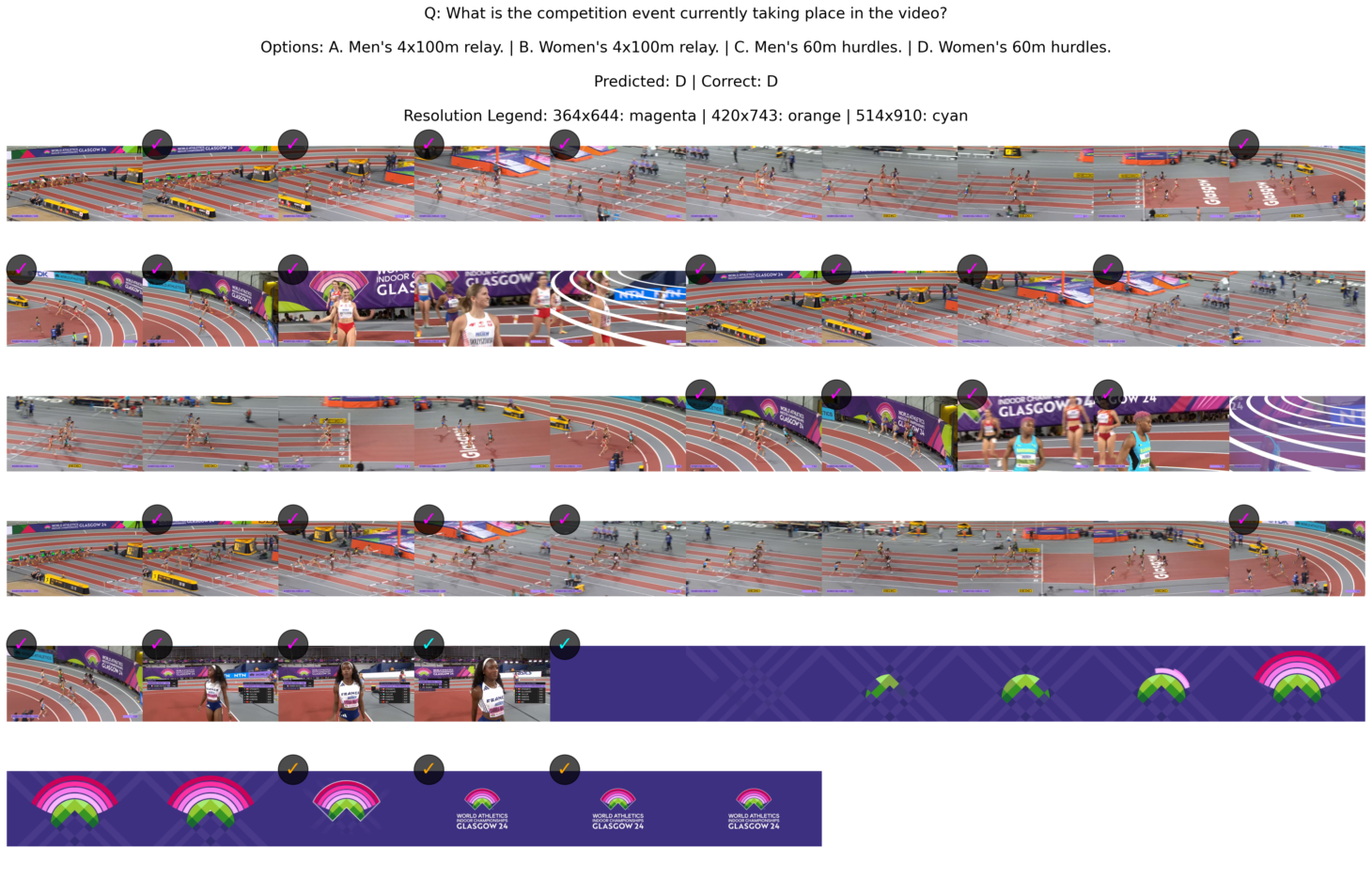}
    \caption{
Visualization of F2C-selected frames for an action-related VQA example (sampled at \textbf{1 FPS}).  
Each clip is assigned an adaptive resolution based on its importance:  
\textcolor{magenta}{364$\times$644} (low resolution),  
\textcolor{orange}{420$\times$743} (medium resolution), and  
\textcolor{cyan}{514$\times$910} (high resolution).  
The figure shows that F2C selects frames covering the full temporal progression of the race—from the start, mid-race dynamics, to the finish—allowing the model to correctly identify the event (\textbf{Women's 60m hurdles}) by preserving both semantic cues and motion continuity.
}
    \label{fig:vis_example1}
\end{figure*}

\begin{figure*}[t]
    \centering
    \includegraphics[width=\linewidth]{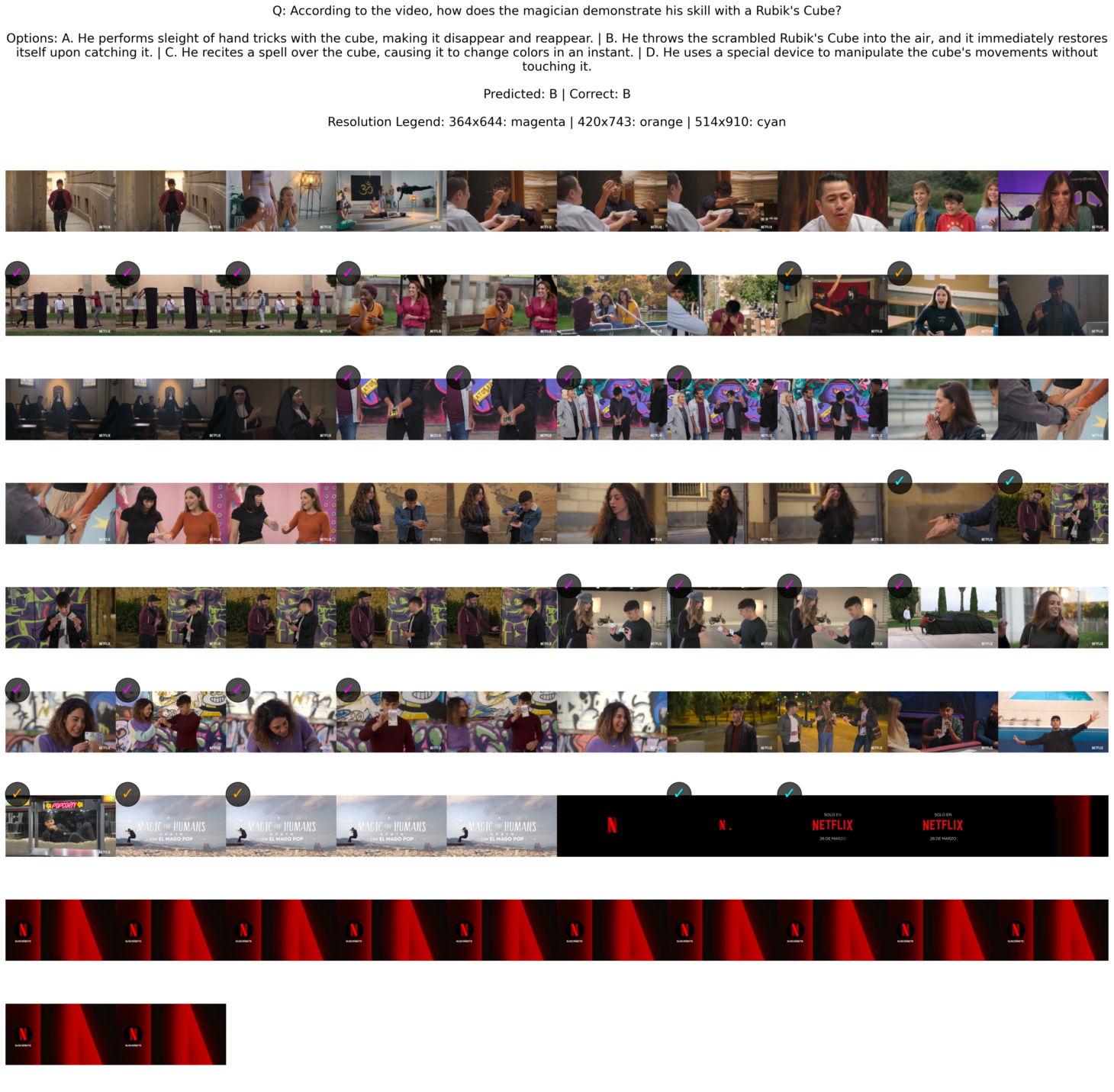}
\caption{\textbf{F2C adaptively balances spatial resolution and temporal coverage.}
Frames are sampled at \textbf{1 FPS}. 
Lower resolution (\textcolor{magenta}{364$\times$644}) is used when a \emph{longer clip length} is needed to capture more temporal context, 
medium resolution (\textcolor{orange}{420$\times$743}) provides a moderate trade-off, 
and higher resolution (\textcolor{cyan}{514$\times$910}) is allocated to frames requiring fine-grained spatial details. 
This adaptive strategy preserves both motion cues and appearance information, enabling the model to correctly infer how the magician demonstrates his Rubik's Cube skill (Predicted: B, Correct: B).}
    \label{fig:vis_example2}
\end{figure*}

\begin{figure*}[t]
    \centering
    \includegraphics[width=\linewidth]{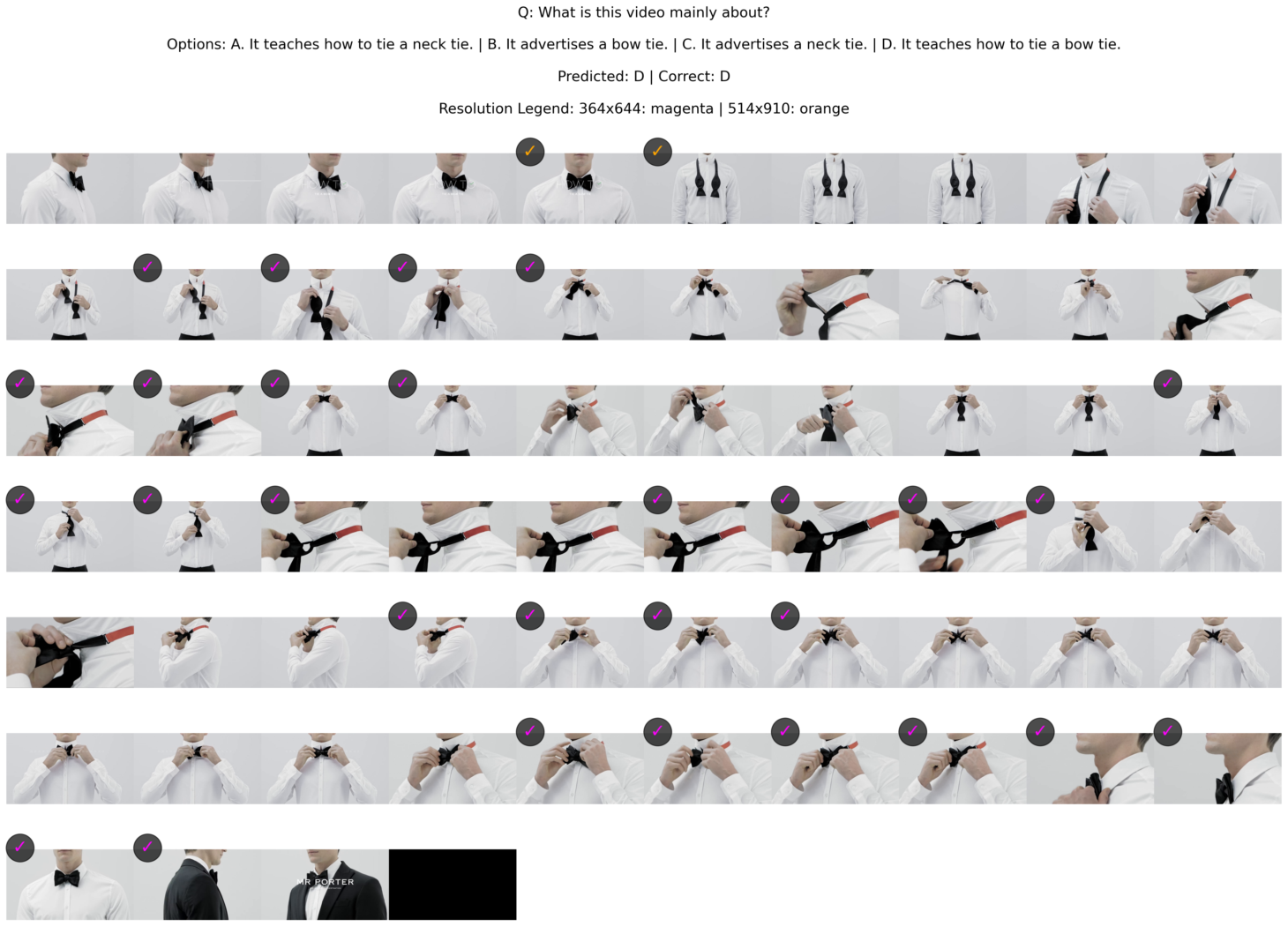}
\caption{
\textbf{F2C on high-level contextual understanding.}
Frames are sampled at 1\,fps and visualized with their selected resolution 
(\textcolor{magenta}{\textbf{364$\times$644}} and 
\textcolor{orange}{\textbf{514$\times$910}}). 
F2C captures the full progression of tying a bow tie—from preparation to the completed knot—demonstrating that the method can also handle questions requiring holistic, high-level understanding rather than only moment-specific actions. 
}
    \label{fig:vis_example3}
\end{figure*}

Figure~\ref{fig:vis_example1} illustrates that F2C effectively captures key frames with coherent temporal progression, which is particularly beneficial for action-related VQA. Figure~\ref{fig:vis_example2} demonstrates F2C's ability to adaptively balance clip length and spatial resolution based on content, while Figure~\ref{fig:vis_example3} shows that the method also performs well on questions requiring high-level contextual understanding. 

\begin{figure*}[t]
    \centering
    \includegraphics[width=\linewidth]{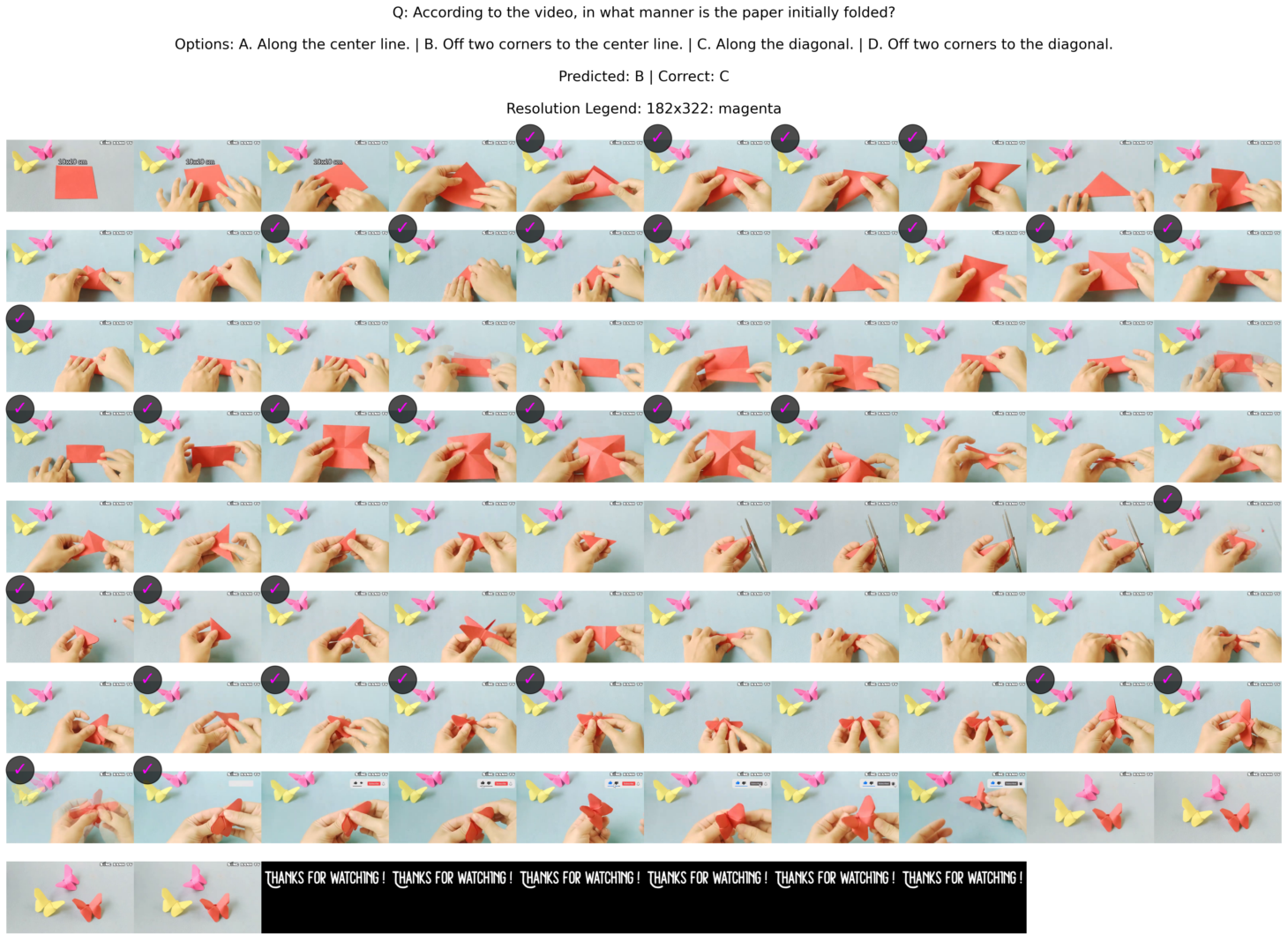}
\caption{\textbf{Failure case in Video-MME.} Although F2C successfully selects frames highly relevant to the query, the downstream VLM fails to perform the required reasoning, resulting in an incorrect prediction. This example illustrates that overall performance remains bounded by the reasoning capability of the underlying VLM, even when key visual evidence is correctly captured.}
    \label{fig:wrong1}
\end{figure*}

\begin{figure*}[t]
    \centering
    \includegraphics[width=\linewidth]{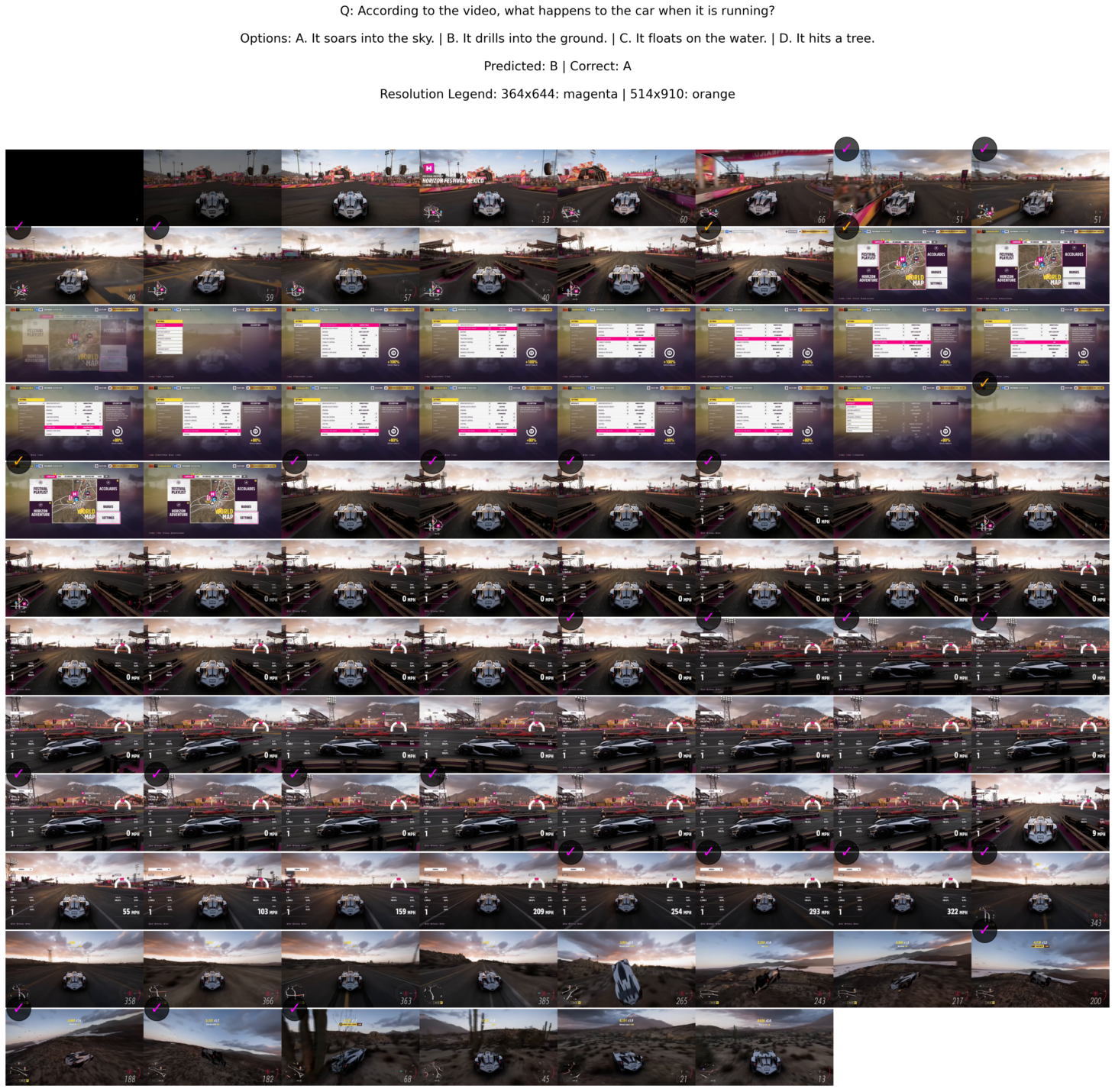}
\caption{\textbf{Failure Case: Key Evidence Not Fully Captured.}
In this example, the critical event appears only briefly within the long video, and F2C does not select the decisive frames required to answer the question correctly. Although many relevant clips are included, the key moment is too sparse and subtle, highlighting a limitation of clip-based selection when essential visual cues are extremely short-lived.}
    \label{fig:wrong2}
\end{figure*}

We additionally present representative failure cases. In Figure~\ref{fig:wrong1}, F2C successfully identifies relevant frames, but the downstream VLM fails to reason correctly, leading to an incorrect answer. In Figure~\ref{fig:wrong2}, for more complex reasoning-oriented questions, F2C does not fully capture the critical event, highlighting potential limitations when the key evidence is extremely sparse or subtle.

\section{More experiments}
\subsection{Impact of the Version of CLIP model}\label{app:clip}
We observe that the choice of CLIP backbone has only a marginal impact on performance. As shown in Table~\ref{tab:clip_versions}, all versions achieve similar accuracies across different frame budgets (8, 16, and 32). While CLIP slightly outperforms others at 16 and 32 frames, and SigLIP leads under the 8-frame setting, the overall differences remain within a narrow margin (typically less than 2\%). This suggests that the improvements brought by our method are robust to the underlying vision-language encoder choice, and the effect of CLIP version is not a dominant factor in determining downstream performance. In our experiments, we adopt SigLIP2 as the default backbone since it offers competitive accuracy while being more efficient at shorter sequence lengths, and it represents a more recent CLIP variant.

\begin{table}[t]
\caption{Comparison of different CLIP versions on Video-MME overall accuracy (\%). Columns show performance under 8, 16, and 32 frames.}
\label{tab:clip_versions}
\centering
\small
\begin{tabular}{lcccc}
\toprule
\textbf{CLIP-version} & \textbf{Max length} & \textbf{8} & \textbf{16} & \textbf{32} \\
\midrule
LongCLIP~\citep{zhang2024longclip}      & 248 & 61.9 & 63.1 & 65.1 \\
CLIP~\citep{radford2021learning}   & 77  & 61.9 & 65.1 & 65.7 \\
SigLIP~\citep{zhai2023sigmoid}        & 64  & 62.5 & 64.8 & 65.4 \\
SigLIP2~\citep{tschannen2025siglip2multilingualvisionlanguage}       & 64  & 61.4 & 63.4 & 65.6 \\
\bottomrule
\end{tabular}

\end{table}

\end{document}